\documentclass[sigconf]{acmart}
\usepackage{amsmath}
\usepackage{pifont}

\usepackage{balance}

\usepackage{epigraph,booktabs}
\usepackage{latexsym}

\usepackage[utf8]{inputenc}

\usepackage{microtype}

\usepackage{graphicx}
\usepackage[font=small]{caption}

\usepackage{color,soul}
\usepackage{tabstackengine}
\usepackage{booktabs}

\newcommand{\figref}[1]{Fig.~\ref{#1}}
\newcommand{\tabref}[1]{Tab.~\ref{#1}}
\newcommand{\secref}[1]{Sec.~\ref{#1}}

\newcommand{\equref}[1]{Eq.~(\ref{#1})}

\newcommand{\beas}{\begin{eqnarray*}}
\newcommand{\eeas}{\end{eqnarray*}}
\newcommand{\bea}{\begin{eqnarray}}
\newcommand{\eea}{\end{eqnarray}}
\newcommand{\bes}{\begin{equation*}}
\newcommand{\ees}{\end{equation*}}
\newcommand{\be}{\begin{equation}}
\newcommand{\ee}{\end{equation}}

\newcommand{\cY}{{\mathcal{Y}}}

\makeatletter
\usepackage{xspace}
\usepackage{subcaption}
\def\@onedot{\ifx\@let@token.\else.\null\fi\xspace}
\DeclareRobustCommand\onedot{\futurelet\@let@token\@onedot}

\def\eg{\emph{e.g}\onedot} \def\Eg{\emph{E.g}\onedot}
\def\ie{\emph{i.e}\onedot} 
 
\def\etc{\emph{etc}\onedot} \def\vs{\emph{vs}\onedot}

\AtBeginDocument{%
  \providecommand\BibTeX{{%
    \normalfont B\kern-0.5em{\scshape i\kern-0.25em b}\kern-0.8em\TeX}}}

\setcopyright{acmcopyright}
\copyrightyear{2022}
\acmYear{2022}
\setcopyright{acmcopyright}\acmConference[MM '22]{Proceedings of the 30th ACM International Conference on Multimedia}{October 10--14, 2022}{Lisboa, Portugal}
\acmBooktitle{Proceedings of the 30th ACM International Conference on Multimedia (MM '22), October 10--14, 2022, Lisboa, Portugal}
\acmPrice{15.00}
\acmDOI{10.1145/3503161.3548161}
\acmISBN{978-1-4503-9203-7/22/10}

\begin{document}

\title{Ordered Attention for Coherent Visual Storytelling}


\author{Tom Braude}
\affiliation{%
  \institution{Reichman University \& Microsoft}
  \city{Hertzliya}
  \country{Israel}}
\email{tom.braude@post.idc.ac.il}

\author{Idan Schwartz}
\affiliation{%
  \institution{Technion}
  \city{Haifa}
  \country{Israel}
}
\email{idandc@technion.ac.il}

\author{Alexander Schwing}
\affiliation{%
 \institution{University of Illinois at Urbana-Champaign}
 \city{Champaign}
 \state{Illinois}
 \country{USA}}
\email{aschwing@illinois.edu}
 
\author{Ariel Shamir}
\affiliation{%
  \institution{Reichman University}
  \city{Hertzliya}
  \country{Israel}}
\email{arik@idc.ac.il}

\renewcommand{\shortauthors}{Tom Braude, Idan Schwartz, Alexander Schwing,  \& Ariel Shamir}

\begin{abstract}
We address the problem of visual storytelling, \ie, generating a story for a given sequence of images. While each story sentence should describe a corresponding image, a coherent story also needs to be consistent and relate to both future and past images. 
Current approaches encode images independently, disregarding relations between images.  Our approach learns to encode images with different interactions based on the story position (\ie, past image or future image). To this end,  we develop a novel message-passing-like algorithm for ordered image attention (OIA) that collects interactions across all the images in the sequence. Finally, to generate the story's sentences, a second attention mechanism picks the important image attention vectors with an Image-Sentence Attention (ISA). The obtained results improve the METEOR score on the VIST dataset by 1\%. Furthermore, a thorough human study confirms improvements and demonstrates that order-based interactions significantly improve coherency (64.20\% \vs 28.70\%). Source code available at \url{https://github.com/tomateb/OIAVist.git}
\end{abstract}


\begin{teaserfigure}
  \centering
    \includegraphics[width=1\linewidth]{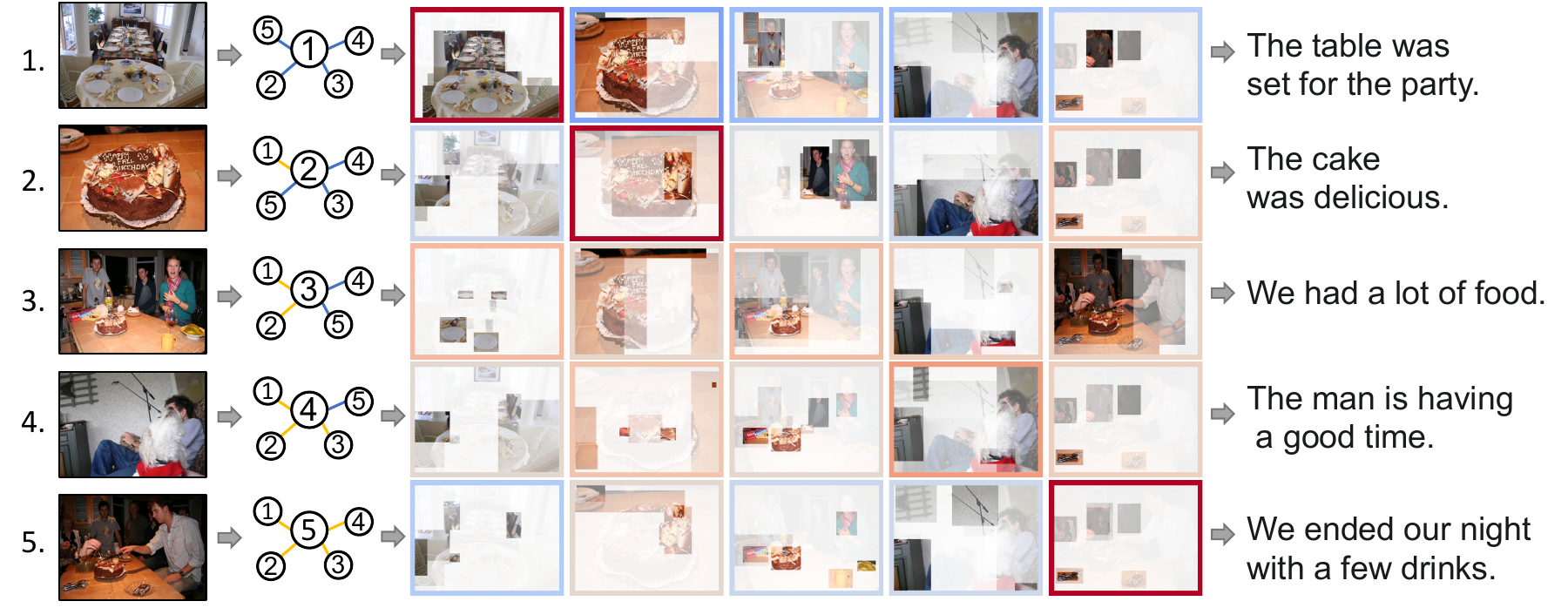}
    \vspace{-.8cm}
    \caption{To promote story coherence with consistent grounding across ordered images, we propose Ordered Image Attention (OIA). By collecting directional interactions, we can identify significant objects throughout the story. In total, five attention maps are calculated, one for each image. The border of attended images indicates how important an image is according to the Image-Sentence Attention (ISA). 
}
    \label{fig:teaser}
\end{teaserfigure}

\maketitle

\section{Introduction}

\textit{Visual Storytelling} (VST)~\cite{Park2015ExpressingAI,Huang2016VisualS} -- the task of generating a story based on a sequence of images -- goes beyond a basic understanding of visual scenes and can be applied in many real-world scenarios, \eg, to support the visually impaired. Moreover, VST reflects on the creative ability of intelligent systems. Although similar in concept to other  cognitive tasks such as image captioning and visual question answering, VST differs as it requires to reason over a \emph{sequence} of images while simultaneously ensuring coherence across multiple generated sentences.
To achieve this, VST methods need to address two major challenges: the first is visual and relates to grounding the story's text to the images. The second is linguistic and relates to the quality of the story. 
Both challenges can be described in terms of coherency: the story should be coherent by itself, and consistent with the images. 

Prior research on VST started to address the aforementioned challenges. Early works expand captioning~\cite{Vinyals2014ShowAT,Xu2015ShowAA,Chen2015MindsEA}, focusing  sentence generation mainly on the current image.
This limits the ability to create a narrative that includes consistent visual information. Prior work also makes limited use of temporal dependence and history, \eg,  sentences that have already been generated are not used. 
Consequently, the output prone to linguistic errors such as \emph{repetitiveness}~\cite{Modi2019TheSR}. To mitigate these issues, later works strive to generate more meaningful stories via adversarial and reinforcement learning~\cite{Wang2018NoMA,Huang2018HierarchicallySR}, which remain delicate to train.

Importantly, images are not independent. 
For example, if the first image in a sequence shows a protest,  the model may want to focus on  signs 
in later images. Conversely, if the last image shows a ring on a finger, then the model should pay attention to wedding-related objects and activities in the preceding images. This is important for VST because sentences are created per image but are part of a story. Hence, objects that the model is focusing on in one image should be conditioned on the selection in other images.

To do this we develop a novel model which (1) implicitly reasons over objects, activities, and their temporal dependencies in each image; and which (2)  improves the coherency of the narrative. 
To reason over objects and activities in each image, \ie, to understand their dependencies and their temporal ordering,  we introduce \emph{ordered image attention} (OIA). 
As illustrated in \figref{fig:teaser}, for each image, OIA accumulates representation information from objects detected within the corresponding image into an attended image representation. Importantly, accumulation factors depend on whether the image precedes or succeeds the image for which we are currently generating the sentence, which permits to establish an order. The attended image representations are subsequently summarized into a context embedding via an Image-Sentence Attention (ISA) unit, before being used for sentence decoding. 

In addition, to alleviate common linguistic mistakes like repetitiveness and to promote coherence in the story, we incorporate information from the story generated up to the current sentence into the sentence generation decoder.  
Specifically, the decoding strategy decays the probability of a word if it has already been used in the story. The decoder also maintains a separate prior over the output probability distribution, independent from the language generation unit. This prior is based on counts of the words that were already predicted in the story. Both the prior, and the Recurrent Neural Net (RNN) decoder output are combined to predict the next word in the sentence. 

Empirical results on the challenging VIST dataset demonstrate that the proposed method generates stories with an improved narrative quality. The method outperforms prior state-of-the-art by 1\% on the METEOR score. Examples of stories generated by the approach are shown in~\figref{fig:teaser}. We also present a user study demonstrating the advantage of the model in terms of coherency (64.20\% \vs 28.70\%).

\section{Related Work}
Vision+Language has been an active area of research for many years, addressing tasks such as image/video captioning, paragraph generation, and visual question answering. 
We briefly review those related areas in the following.
\begin{figure*}[t]
\centering
\includegraphics[width=0.97\linewidth]{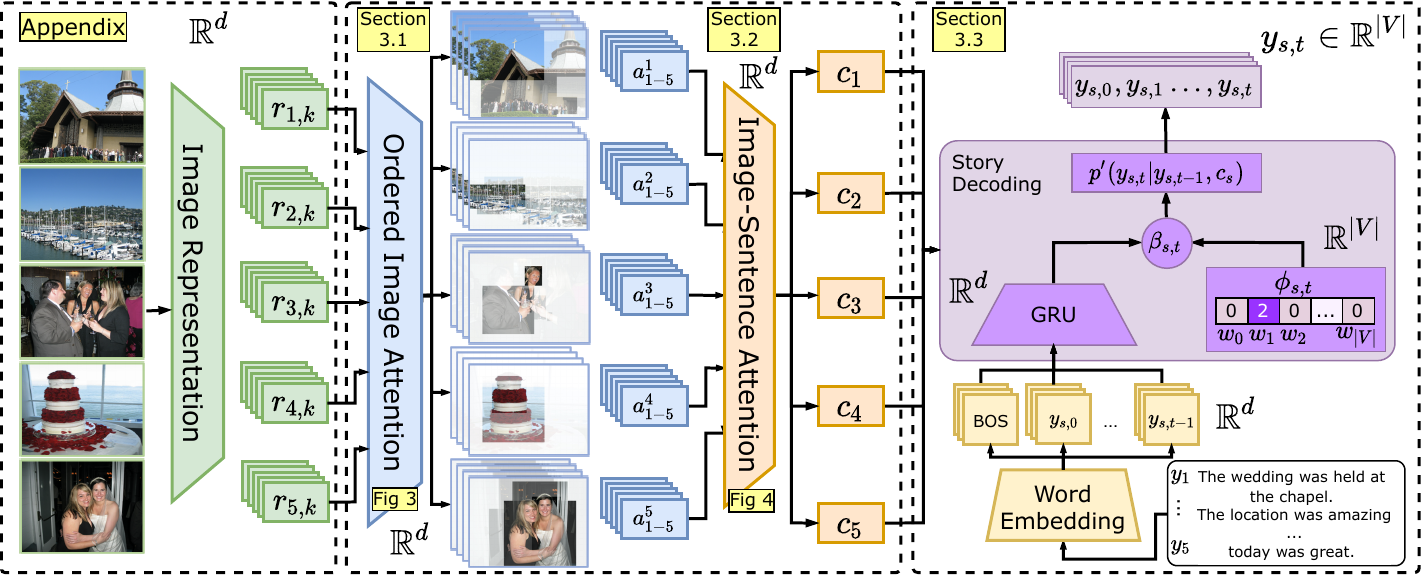}
\vspace{-0.2cm}
\caption{Our 
architecture for Visual Storytelling synthesis.
}
\vspace{-0.5cm}
\label{fig:fullModel}
\end{figure*}   

\subsection{Visual Storytelling} 
\citet{Huang2016VisualS} introduce the visual storytelling task. Visual storytelling is similar to captioning. Thus, early methods adapted captioning mechanisms, introducing context between story sentences~\cite{GonzalezRico2018ContextualizeSA}. Following, \citet{Kim2018GLAC} used a seq2seq~\cite{Sutskever2014SequenceTS} approach built on a decoding sampling strategy to reduce repetition. Here, we use a dynamic data-driven approach where each word is penalized differently based on its average count. Next, \citet{Wang2018NoMA}  discuss the difficulty of learning stories with imaginary details that do not appear in the imagery. To that end, an adversarial reward system is used to improve the output stories. Several works use a reinforcement learning approach based on the interrelationships between images~\cite{Huang2018HierarchicallySR}. Recently, state-of-the-art results were obtained by generating scene graphs for each image in the sequence~\cite{Wang2019StorytellingFA}. Following, \citet{Li2019InformativeVS} and \citet{Zhang2020VisualSV} rely on preprocessing to ground visual elements. \citet{Yang2019KnowledgeableSA} and \citet{hsu2020knowledge} enrich the data with an external word common-sense knowledge graph.\citet{Wang2019StorytellingFA} model relations within the image with scene graphs, which requires expensive annotations and \citet{qi2021latent} modify the Transformer architecture to model inter region relations.  \citet{hong2020diverse} also uses scene graphs with global embeddings to achieve coherence. Instead, our model forms ordered attention, which permits the focus on consistent objects without additional supervision. More recent approaches~\cite{yu2021transitional, fan2021visual, su2021bert} employ large pre-trained models for visual storytelling.  In our method, we do not rely on large-scale external knowledge.

\subsection{Image Captioning}
\citet{Barnard2003MatchingWA} first explored annotating images with text. Since then, image/video captioning has seen a surge of research activity. Initial work utilized pre-trained image embeddings from a CNN network. The success of attention mechanisms for language translation quickly transferred to image captioning as well~\cite{Xu2015ShowAA}. Later work leveraged advances in object detection and proposed a bottom-up/top-down attention approach to attend to specific objects in the image instead of fixed spatial regions~\cite{Anderson2017BottomUpAT}. Further improvements include modeling spatial and semantic interactions with graph neural networks and transformers~\cite{pan2020x,dosovitskiy2021image,guo2020normalized}. A key factor for improving model performance is pre-training on large amounts of data and fine-tuning on curated supervised datasets~\cite{zhang2021vinvl,devlin2018bert,li2020oscar}. Recently, multimodal pre-training models like CLIP are enabling large-scale pre-trained LMs to be guided in a zero-shot fashion~\cite{tewel2021zero}.  As opposed to captioning images, visual storytelling aims to ground a visual story from multiple cues. This means that only objects related to the narrative should be highlighted, which is the focus of our work.

\subsection{Multimodal Attention}
Multimodal problems are characterized by input data that comes from different domains, \eg, visual and linguistic. This raises two challenges: 1) how to model interactions between different domains, and 2) how to manage the large input data. Considering those challenges, attention has been a prominent tool as it models  interactions to select the important elements. In early work, \citet{Xu2015ShowAA} used interaction-based attention with the image at each caption generation step. This idea was later extended to visual question answering \cite{xu2016ask}. To imitate multi-step reasoning, \citet{yang2015stacked} stacked attention modules sequentially. Later, many works concentrated on better vector-fusion modeling \cite{fukui2016multimodal,kim2016hadamard,ben2017mutan,yu2018beyond}. Importantly, \citet{lu2016hierarchical}  suggested attending to the visual and textual modalities separately. Afterward, \citet{kim2018bilinear} proposed a bilinear module that efficiently generates attention for every pair. Following \citet{lu2016hierarchical}, \citet{Schwartz2017HighOrderAM,Schwartz2019FactorGA} suggested a general framework that extends attention to any number of utilities via local and interaction-based factors. We improve upon those ideas by suggesting an ordered attention. This ensures that interaction modeling is affected by the image position in a sequence.

\section{Method}
The goal of visual storytelling is to generate a story, composed of $N$ \emph{ordered} sentences $\{y_s | 1\leq s \leq N\}$, given an \emph{ordered} sequence of images $I = \{I_s | 1\leq s \leq N\}$. Each sentence $y_s = (y_{s,0}, \dots, y_{s,t}, \dots)$ is composed of words $y_{s,t}\in\cY$ from vocabulary $\cY$.

The order in which the images are given is essential as it defines the plot line of the story. The story should be focused, \ie, each sentence should be related to the remainder of the story. Importantly, the sentences should form a coherent body of text describing the set of images, and not only a set of related information. For instance, the story \textit{``The church was beautiful. The bride and groom walk down the aisle. The cake was amazing.''} is less coherent than: \textit{``We went to the church for the wedding today. The bride and groom were excited for the day. Both cut the cake together.''} 

\noindent\textbf{Overview:} To address this challenge, we develop the model illustrated in \figref{fig:fullModel}. It infers conditional probabilities $p'(y_{s,t}|y_{s,t-1},c_s)$ for the $t$-th word $y_{s,t} \in \cY$  in sentence $y_s$ given the previous word $y_{s,t-1}$ and the context embedding $c_s$ for sentence $s$. The context embedding $c_s$ summarizes  region representations $r_{i,k}$ of all $K$ object regions across all $N$ images $I_i$ ($i\in[1,N]$, $k\in[1,K]$) via Ordered Image Attention (OIA) (Sec.~\ref{sec:oia}) and Image-Sentence Attention (ISA) (Sec.~\ref{sec:sentence attention}). Specifically, when generating sentence $s$, OIA computes an attended image representation $a_i^s$ for every image $I_i$ by attending to the $K$ region representations $r_{i,k}$ (Sec.~\ref{sec:oia}). These attended image representations $a_i^s$ are subsequently summarized into the context embedding $c_s$ via an image-sentence attention (Sec.~\ref{sec:sentence attention}).

Below we first discuss computation of the attended image representation $a_i^s$  (Sec.~\ref{sec:oia}), before detailing computation of the context embedding $c_s$ (Sec.~\ref{sec:sentence attention}) and computation of the conditional probabilities $p'(y_{s,t}|y_{s,t-1},c_s)$ (Sec.~\ref{sec:decoder}).

%
Ordered Image Attention (OIA)  is designed to 1) form a structure across ordered images and to 2) select the relevant objects per image. For this we model preceding and subsequent interactions separately  using different attention factors. We calibrate each factor's importance with trainable scalars, which forms a graph of dependencies between the images. For each sequence of $N$ images, the model infers a total of $N^2$ attention maps, one per image for each sentence.  We  detail this  module next.

\subsection{Ordered Image Attention (OIA)}

\subsubsection{Attention Belief}

For each image $I_i = \{r_{i,1}, \ldots r_{i,K}\}$ we consider a set of $K$ regions, represented by their feature vectors $r_{i,k}\in\mathbb{R}^d$, where $d$ is the objects' embedding dimension. Suppose we are currently generating sentence $y_s$  ($1\leq s \leq N$). To do this we first compute an attended image representation $a_i^s$ as follows
\begin{gather}
  a_i^s = \sum_{k=1}^K b_{i,k}^{s} r_{i,k}, 
\end{gather}
where $b_{i, k}^{s}\geq 0$  is the attention belief highlighting the importance of the $k$-th object in the $i$-th image when generating the $s$-th sentence. Importantly, for every image $I_i$ we require $b_{i,k}^{s}$ to be a valid probability distribution, \ie, we also enforce   $\sum_{k=1}^K b_{i,k}^{s} = 1$ $\forall s,i$. 

\begin{figure}[t]
	\centering
    \includegraphics[width=1\linewidth]{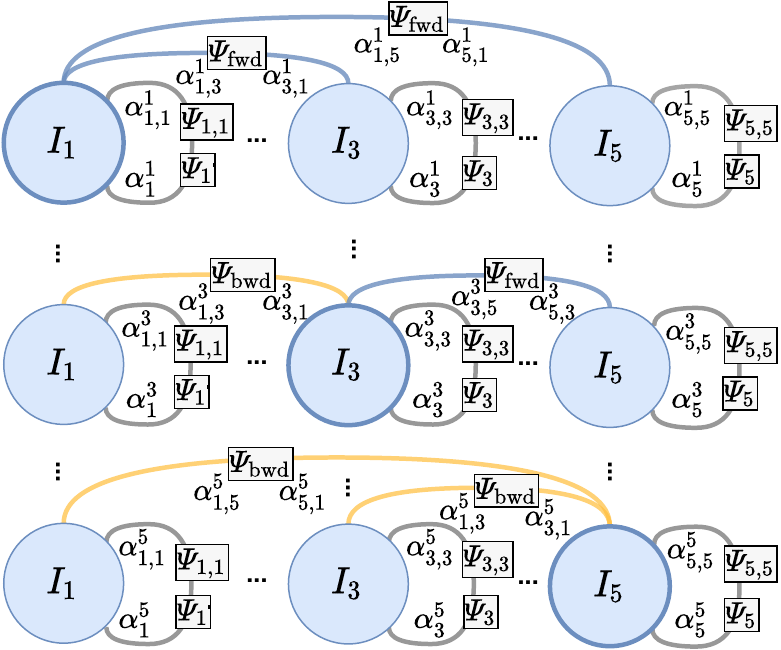}
    \caption{Illustration of Ordered Image Attention. Each node represents an image attention belief. For each sentence, we connect all the images with the sentence-corresponding image. The relative position to this image determines whether the connection is modeled with the  $\Psi_{\text{bwd}}$ factor (for preceding images) or the $\Psi_{\text{fwd}}$  factor (for subsequent images; see Eqs.~[\ref{eq:p_f}-\ref{eq:self}]).  We infer the attention belief by collecting interactions and local object information within the image (see Eqs.~[\ref{eq:b_a}-\ref{eq:b_c}]). We use scalars to calibrate the importance of each factor. In total, we generate 25 attention maps, one per image for every sentence. }
    \vspace{-.5cm}
    \label{fig:dafga}
\end{figure}

The object attention belief $b_{i,k}^s$ is dependent on all the input data, \ie, other objects and images. To avoid complex computation, we factorize the belief $b_{i,k}^s$ into two pairwise dependencies that preserve the order, and a local term. For the pairwise terms we use $\mu_{j\to i}^{\text{bwd}}$, which is a message from a preceding image $I_j$, or $ \mu_{j\to i}^{\text{fwd}}$, which is a message from a subsequent image $I_j$. We also use $ \mu_{i\to i}$ for self-messages. 
Additionally, we include a local factor $\Psi_{i}(r_{i,k})$ that considers the object representation. Unlike the messages mentioned before, the local factor does not rely on interactions with other objects.
We aggregate all the messages along with the local factor as illustrated in  \figref{fig:dafga}. For normalization we employ  a softmax. 
Formally we compute the attention belief $b_{i,k}^s$ by distinguishing three cases. If $i=s$ we have
\begin{eqnarray}\label{eq:b_a}
b_{i,k}^s &\propto& \exp ( \alpha_{i}^{s} \Psi_{i}(r_{i,k})  + \alpha_{i,i}^{s} \mu_{i\to i} (r_{i,k})+\\
\nonumber&&\sum_{j<i} \alpha_{i,j}^{s} \mu_{j\to i}^{\text{bwd}}(r_{i,k})+\sum_{j>i} \alpha_{i,j}^s \mu_{j\to i}^{\text{fwd}}(r_{i,k})).
\end{eqnarray}
If $i<s$ we use
\begin{eqnarray}\label{eq:b_b}
b_{i,k}^s &\propto&\exp ( \alpha_{i}^{s} \Psi_{i}(r_{i,k})  + \\
\nonumber&&\alpha_{i,i}^{s} \mu_{i\to i}(r_{i,k})+\alpha_{i,s}^{s} \mu_{s\to i}^{\text{bwd}} (r_{i,k})).
\end{eqnarray}
If $i>s$ we obtain
\begin{eqnarray}\label{eq:b_c}
b_{i,k}^s &\propto&\exp ( \alpha_{i}^{s} \Psi_{i}(r_{i,k})  +\\
\nonumber&&\alpha_{i,i}^{s} \mu_{i\to i}(r_{i,k})+ \alpha_{i,s}^{s} \mu_{s\to i}^{\text{fwd}}(r_{i,k})).
\end{eqnarray}
In all three cases $\alpha_{i}^{s}, \alpha_{i,i}^{s},  \alpha_{i,j}^{s} \in \mathbb{R}$ are scalars used to calibrate the importance of different messages for a given sentence. These scalars form a dependency structure between images for each of the generated sentence indices. Intuitively, when we generate the first sentence, the attention belief might depend more on subsequent images, to correctly identify the story event, \eg, a wedding, a parade, \etc. Thus, the scalars will promote interaction with later images. An analysis of these scalars is provided in the appendix. Next, we define the different types of messages. 

\subsubsection{Pairwise Messages and Factors}

A message aggregates interaction scores from an image to an object. The three messages $\mu_{j\to i}^{\text{bwd}}, \mu_{j\to i}^{\text{fwd}}$ and $\mu_{i\to i}(r_{i,k})$ are computed as follows:
\begin{equation}\label{backwardlooking}
  \mu_{j\to i}^{\text{bwd}}(r_{i,k}) = \sum_{k'=1}^K  \Psi_{\text{bwd}}(r_{i,k}, r_{j,k'}),
\end{equation}
\begin{equation}\label{forwardlooking}
  \mu_{j\to i}^{\text{fwd}}(r_{i,k}) =  \sum_{k'=1}^K  \Psi_{\text{fwd}}(r_{i,k}, r_{j,k'}), ~\text{and}
\end{equation}
\begin{equation}
  \label{eq:selg-message}   \mu_{i\to i}(r_{i,k}) = \sum_{k'=1}^K \Psi_{i,i}(r_{i,k}, r_{i,k'}).
\end{equation}
Importantly, these messages collect three different types of order-dependent interaction factors: 
(1) A backward image interaction, namely $\Psi_{\text{bwd}}(r_{i,k}, r_{j,k'})$. This interaction models relations to the preceding $j$-th image in the sequence. 
(2) A forward image interaction, namely $\Psi_{\text{fwd}}(r_{i,k}, r_{j,k'})$.  This interaction models relations to the subsequent $j$-th image in the sequence.  
(3) The self interaction factor, namely $\Psi_{i,i}(r_{i,k},r_{i,k'})$, which takes into account interactions between objects within the image. 
We formally define the different factors next. 

\noindent\textbf{Interaction factors:} 
A commonly used practice to capture interactions across attention mechanisms is to first embed the elements into a joint Euclidean space followed by a dot-product \cite{vaswani2017attention,Schwartz2017HighOrderAM,gao2019dynamic,Schwartz2019FactorGA}. While we follow the same practice, we define three types of interaction factors to preserve the order. Consider  two objects, $r_{i,k}\in I_i$ from the sentence-corresponding image  and $r_{j,k'}\in I_j$ from the interacting image. We describe three types of interactions: for interactions with subsequent images (\ie, $j > i$) we use
\begin{equation}\label{eq:p_f}
  \Psi_{\text{fwd}}(r_{i,k}, r_{j,k'}) \!=\! \left(\frac{L_{\text{fwd}} r_{i,k}}{\|L_{\text{fwd}} r_{i,k}\|_2}\right)^{\!\!\top}\!\!\left(\frac{R_{\text{fwd}} r_{j,k'}}{\|R_{\text{fwd}} r_{j,k'}\|_2}\right).
 \end{equation}
 For interactions with preceding images (\ie, $j < i$) we use 
 \begin{equation}\label{eq:p_b}
 \Psi_{\text{bwd}}(r_{i,k}, r_{j,k'}) \!=\! \left(\!\frac{L_{\text{bwd}} r_{i,k}}{\|L_{\text{bwd}} r_{i,k}\|_2}\!\right)^{\!\!\top}\!\!\left(\!\frac{R_{\text{bwd}} r_{j,k'}}{\|R_{\text{bwd}} r_{j,k'}\|_2}\!\right).   
 \end{equation}
For interactions within the image  (\ie, $j = i$) we have 
\begin{equation}\label{eq:self}
\Psi_{i,i}(r_{i,k}, r_{i,k'}) \!=\! \left(\frac{L_{i,i} r_{i,k}}{\|L_{i,i} r_{i,k}\|_2}\right)^{\!\!\top}\!\!\left(\frac{R_{i,i} r_{i,k'}}{\|R_{i,i} r_{i,k'}\|_2}\right).
\end{equation}
Note, $L_\text{fwd}, R_\text{fwd}, L_\text{bwd}, R_\text{bwd}, L_{i,i}, R_{i,i}\in \mathbb{R}^{d\times d}$ are trainable shared weights across the entire image sequence. Also, the object from the sentence-corresponding image will always be on the left side of the factor equation. Thus, the factor embeddings preserve the order.

\begin{figure}[t]
	\centering
    \includegraphics[width=1\linewidth]{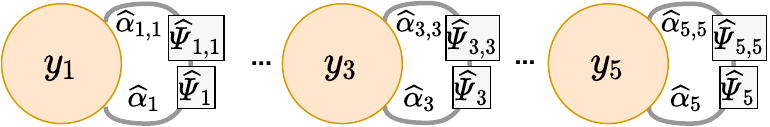}
    \caption{Illustration of  ISA.  The attention selects the attended image representation per sentence. We model interactions between attended images of the same sentence to compute each image's importance. Note, each node represents a sentence attention belief over the attended images.}
    \vspace{-0.5cm}
    \label{fig:sa}
\end{figure}

\noindent\textbf{Local factor:} 
Differently from the previous interactions the following factor captures how important an object is based solely on the object representation. Given an object $r_{i,k}\in I_i$, we define the local factor as, 
    \begin{align}
        \label{eq:local}
        \Psi_i(r_{i,k}) = v^\top \operatorname{ReLU}(V r_{i,k}),
    \end{align}
where $v \in \mathbb{R}^d, V \in \mathbb{R}^{d\times d}$ are trainable weights. 

\begin{table*}[t]
    \centering
    \caption{Quantitative results on the VIST dataset for METEOR, BLEU-1$\ldots$4,  ROUGE-L and CIDEr. The primary metric is METEOR. The `Img Feat' column describes the pretrained image features. All models utilize a ResNet~\cite{He2015DeepRL} backbone except CS\&T which employs an Inception v3 model~\cite{Szegedy2015RethinkingTI}. FC and Spatial refer to features extracted from the penultimate layer and the preceding one accordingly. F-RCNN are based on~\cite{Anderson2017BottomUpAT}.}
    \vspace{-0.3cm}
	\scalebox{1}{
            \begin{tabular}{lcccccccc} 
                \toprule
                Method                    & M     & B-1           & B-2           & B-3           & B-4           & R   & C & Img Feat\\ 
                \midrule
               CS\&T[\citenum{GonzalezRico2018ContextualizeSA}]                    & 34.4    & 60.1          & 36.5          & 21.1          & 12.7         & 29.2 & 7.1          & FC\\
               AREL[\citenum{Wang2018NoMA}]                      & 35.0        & 63.8          & 39.1          & 23.2          & 14.1          & 29.5  & 9.4   & FC\\
               KS[\citenum{Yang2019KnowledgeableSA}] & 35.2     & 66.4          & 39.2          & 23.1          & 12.8          & 29.9 & \textbf{12.1}          & FC\\
                HSRL[\citenum{Huang2018HierarchicallySR}]                     & 35.2                   & -             & -             & -             & 12.3          &  29.5 & 8.4          & Spatial\\
                StoryAnchor[\citenum{Zhang2020VisualSV}]               & 35.5             & 65.1          & 40.0 & 23.4          & 14.0         & 30.0    & 9.9 & FC\\ 
                SGVST[\citenum{Wang2019StorytellingFA}]  & 35.8          & 65.1          & 40.1          & 23.8          & 14.7          &  29.9        & 9.8 & F-RCNN\\
                SGEmb[\citenum{hong2020diverse}] & 35.6          & 62.2 & 38.7 &. 23.5. & 14.8          &  30.2       & 8.6 & F-RCNN\\
                \midrule
                \textbf{Ours}  & \textbf{36.8}$\pm$0.1                     & \textbf{68.4}$\pm$0.7 & \textbf{42.7}$\pm$0.3          & \textbf{25.2}$\pm$0.2            & \textbf{15.3}$\pm$0.2           & \textbf{30.2}$\pm$0.1        & 10.1$\pm$0.2& F-RCNN\\
                \bottomrule
            \end{tabular}
        } 
        \vspace{-0.3cm}
     \label{tab:resultsall}
\end{table*}
\subsection{Image-Sentence Attention (ISA)}
\label{sec:sentence attention}

\label{sec:oia}

In a next step we summarize the attended image representations  $a_i^s$ produced by OIA 
to compute the context embedding $c_s$ for the sentence $s$ that we wish to generate. For this we use the Image-Sentence Attention (ISA) unit. It picks the relevant image context for generating the specific sentence.  
Formally we obtain the context embedding via
\begin{equation}
 c_s = \sum_{i=1}^N \hat b_{s, i} a_i^s,
\end{equation}
where attention factors
\begin{gather} \hat b_{s, i}\propto \exp \left( \hat\alpha_{s} \hat\Psi_{i}(a_i^s) + \hat\alpha_{s,s}\hat\mu_{s\to s}(a_i^s)\right),\end{gather} and where $\hat\alpha_{s}, \hat\alpha_{s,s}\in \mathbb{R}$ are scalars. To avoid spurious correlations between sentences, we consider only  self interactions and a local factor. This is illustrated in \figref{fig:sa}. 
The self-message of the attended image representation $a_i^s$ is 
\begin{gather} \hat\mu_{s\to s}(a_i^s)=\sum_{j=1}^N\hat\Psi(a_i^s, a_j^s).\end{gather}
Finally, the self and local factors are defined with a different set of weights following  \equref{eq:self} and \equref{eq:local} respectively.


\subsection{Story Decoding}
\label{sec:decoder}
The goal at each timestep of decoding is to compute the conditional probability $p(y_{s,t}|y_{s,t-1},c_s)$ where $y_{s,t} \in \cY$  is the $t$-th word in sentence $y_s$,  $\cY$ is the vocabulary and $c_s$ is the context embedding detailed in Sec.~\ref{sec:sentence attention}. 
For this we use a GRU recurrent unit, tasked with generating probabilities over the vocabulary conditioned on the context embedding $c_s$  and the previously generated token $y_{s,t-1}$: $p(y_{s,t}=w|y_{s,t-1}, c_s)\propto$
\begin{eqnarray}
&\exp(\beta_{s,t}\cdot g_{w}(y_{s,t-1}, h_{s,t-1}, c_s) \notag\\
&+ (1-\beta_{s,t})\cdot f_{w}(\phi_{s,t})), \label{word_p}
\end{eqnarray}
where $g_w$ is the output of a GRU unit for the word $w$. We set the GRU hidden dimension to  $d$.  $h_{s,t-1}\in \mathbb{R}^{d}$ is the hidden state at timestep $t-1$ for sentence $s$. $f: \mathbb{R}^{|\cY|}\to \mathbb{R}^{|\cY|}$ is a learned prior over the vocabulary based on a bag-of-words prior histogram $\phi_{s,t}$, which we describe in the next paragraph. The purpose of $f$ is to reduce text repetitions. $f_w$ denotes the value of $f$ for a word $w$. We also incorporate a calibration gate $\beta_{s,t}: \mathbb{R}^d \to [0, 1]$ for functions $f$ and $g$ using
\begingroup
\fontsize{10pt}{10pt}\selectfont
\begin{align}
\beta_{s,t} = \sigma\left(v_{\beta}^\top \operatorname{tanh}(G_g h_{s,t} + G_f W_1(\phi_{s,t}))\right).
\label{eq:gate}
\end{align}
\endgroup
Here, $G_g\in \mathbb{R}^{d \times d}$ and $G_f\in \mathbb{R}^{d \times \gamma}$ are trained projections of the GRU hidden state and the bottleneck layer respectively, $v_{\beta}\in \mathbb{R}^{d}$ are learned weights and $\sigma$ is the sigmoid function. $W_1$ is obtained from the prior as discussed next. 

\noindent\textbf{Bag-of-words (BOW) prior:} 
Remembering history during storytelling permits to stay on topic and advance the story in the desired direction.  Although quite intuitive, mimicking this ability is not trivial. \Eg, most approaches for VST generate all the sentences in parallel. Converting the parallel sentence generation into a sequential one implies a major computational overhead during training. 

To address this, we propose a simple yet effective learnable framework that does not require sequential training while still exploiting information found in prior sentences. The history is represented via a bag-of-words histogram $\phi_{s,t}$, which includes all words that have been used until timestep $t$ for the $s$-th sentence. During training, we initialize $\phi_{s,t=0}$ with the ground truth history counts found in the previous $s-1$ sentences. We update the statistics at each timestep with the predicted word $y_{s',t}$ for $s'<s$, and produce the next state of the counter $\phi_{s,t+1}$. At inference we generate sentences sequentially and update $\phi_{s,t}$ with the predicted words. $\phi_{s,t}$ is fed through a shallow bottleneck network to obtain the prior $f$, composed of two layers $W_1\in \mathbb{R}^{\gamma \times |\cY|}$ and $W_2\in \mathbb{R}^{|\cY| \times \gamma}$ without activation, where $\gamma$ is the bottleneck dimension:
\begin{align}\label{eq:bottleneck}
f(\phi_{s,t}) = W_2(W_1(\phi_{s,t})).
\end{align}
Also note the use of $W_1(\phi_{s,t})$ in the gate (\equref{eq:gate}).

\noindent\textbf{Intra-repetition regularization:} To regularize intra-repetitions, we 
decay the probability of previously used words during sentence generation. A critical aspect of this approach is to exclude words that appear frequently in the language (\eg, was, were, am). 
 For this we pre-process the training set to calculate the average story frequency $\rho({w})$ of a word $w$ via $\rho({w}) = \frac{\#\ \text{appearances of word~} w}{\#\ \text{stories~} w \text{~was used}}$. The final count for word $w$ at timestep $t$ is calculated as  $\phi_{s,t}'(w) = \text{max}[0,(\phi_{s,t}(w)-\rho(w) + 1)] $.  
Intuitively, a word will not be penalized before it is used more than the prior belief average $\rho(w)$. The final probability for word $w$ being used is given by 
\begingroup
\fontsize{9.5pt}{9.5pt}\selectfont
\begin{align}\label{intra_rep_eq}
p'(y_{s,t}=w|y_{s,t-1}, c_s) = \frac{p(y_{s,t}=w|y_{s,t-1}, c_s)}{\pi\cdot \phi_{s,t}'(w) + 1},
\end{align}
\endgroup
where $\pi \geq 0$ is a constant hyper-parameter. A penalty of 2 proved to work best on the validation set.



\section{Results}
\label{sec:res}
\noindent\textbf{Dataset: }To train and test the model we use the VIST dataset.  All images were collected from Flickr albums. All images from a story were taken from the same album.  Each image sequence has five reference stories. Approximately 2.5 of the stories are based on human annotations, while the rest are rewrites. The overall numbers are 40,098 training stories, 4,988 validation stories, and 5,050 test stories.  

\noindent\textbf{Image Representation:}
\label{sec:IREP}
An initial pre-processing step represents each of the input images $I_i$ via $K$ regional features $r_{i,k}\in \mathbb{R}^d, 1\leq k \leq K$. For this we use bottom-up attention features~\cite{Anderson2017BottomUpAT}. Specifically, for each image $I_i$ we first extract the top $K$ region features $e_{i,k} \in \mathbb{R}^{m}$. Hereby, $e_{i,k}$ is an $m$-dimensional feature vector extracted from a pre-trained image classification network~\cite{He2015DeepRL} along with their respective bounding boxes $b_{i,k} \in \mathbb{R}^4$, and classes $c_{i,k}\in \mathbb{N}$.  
The final $d$-dimensional representation $r_{i,k} \in \mathbb{R}^d$, of each region is defined by a combination of the extracted semantic features. Formally,
\begin{align}
r_{i,k} = W_r[W_e e_{i,k} + W_b b_{i,k} + E_c(c_{i,k})], 
\end{align}
where $W_r\in \mathbb{R}^{d\times d}$, $W_e \in \mathbb{R}^{d\times m}$, $W_b\in \mathbb{R}^{d\times 4}$, and $E_c$ are trainable parameters shared between all images. We set $K=36$ in our proposed model. Biases and normalization are omitted for readability.

\noindent\textbf{Evaluation metrics:} 
As suggested by the creators of VIST, METEOR~\cite{Banerjee2005METEORAA} correlates best with human judgement.  We also report BLEU~\cite{Papineni2001BleuAM}, ROUGE~\cite{Lin2004ROUGEAP}, and CIDEr~\cite{Vedantam2014CIDErCI} and compare to prior work where available.  The metrics are based on word correspondence with human references, which is unsuitable for measuring visual storytelling quantities such as coherence. For example, the ROUGE and CIDEr scores are almost identical for all the recent years' baselines. While our experiments indicate statistically significant improvements across all metrics, we emphasize that human evaluation are currently the most reliable way to evaluate visual storytelling approaches.


\subsection{Quantitative analysis}\label{experiments}

\noindent\textbf{Comparison to state-of-the-art:} 
In \tabref{tab:resultsall} we compare the method to recent baselines. Early methods did not take into account visual-spatial information (\ie, they employed FC features), which harms the performance (\eg, 35.5\% \vs 36.8\% on METEOR). \citet{Wang2019StorytellingFA} utilize image representations similar to our approach but do not consider relations between different images, resulting in a 1\% drop on METEOR, showing that  ordered structure encoding with OIA is beneficial. 
SGVST and StoryAnchor map images to distinct topics based on external knowledge. On the other hand, our approach is trained end-to-end. Furthermore, our image representations depend on all the images in a sequence. In contrast, SGVST uses scene graphs. Such models are pre-trained with an external model for generating scene graphs.  Finally, ~\citet{Yang2019KnowledgeableSA}  enhance the input with an external commonsense dataset. CIDEr scores are significantly higher, yet this improvement is not reflected in all metrics. 
Our work improves the state-of-the-art METEOR score from 35.8\% to 36.8\%. This increase is larger than the
0.8 increase of all advances since the 2018 VIST challenge (\ie, 35.0\% \vs 35.8\%). 

\begin{figure}[t]
	\centering
    \begin{subfigure}[t]{1\linewidth}
    	\centering
    	\includegraphics[width=0.7\linewidth]{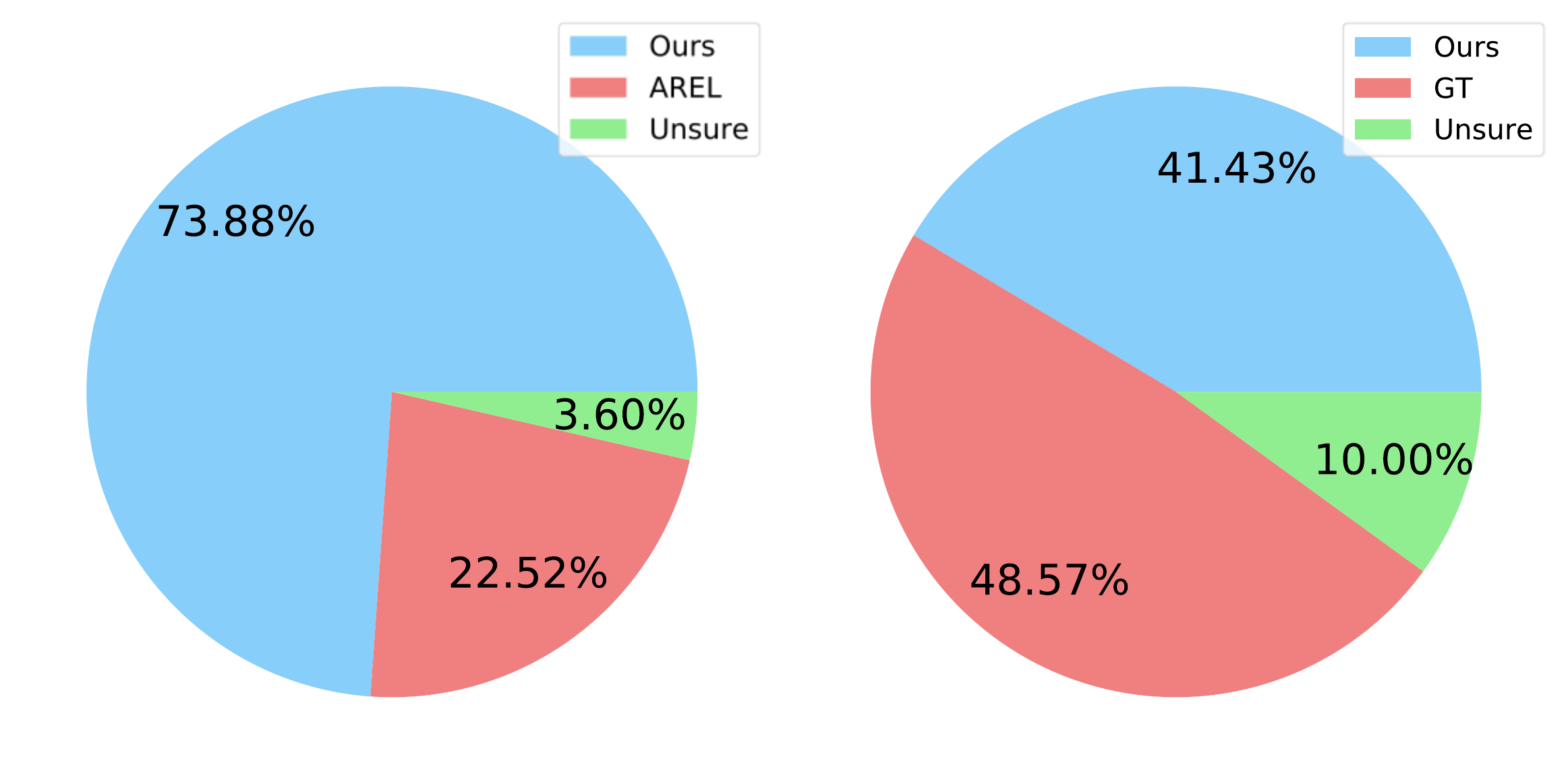}
    	\vspace{-0.3cm}
    	\caption{Human-like property comparison.}
    	\label{fig:comparative}
    \end{subfigure}
    \vfill
    \begin{subfigure}[t]{1\linewidth}
    	\centering
    	\includegraphics[width=0.7\linewidth]{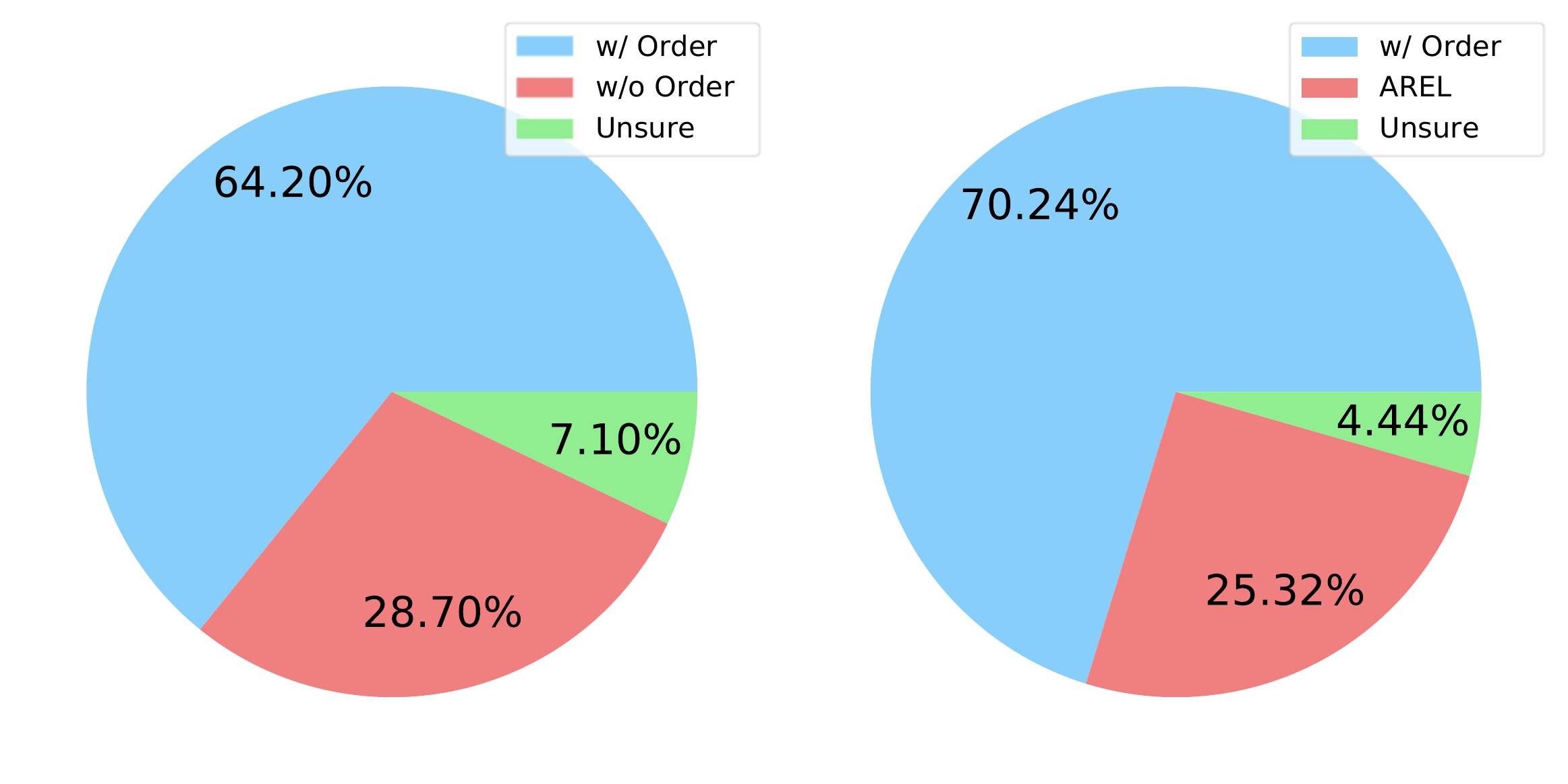}
    	\vspace{-0.3cm}
    	\caption{Coherence property comparison.}
    	\vspace{-0.3cm}
    	\label{fig:dir_compare_qual}
    \end{subfigure}
    \caption{Human evaluation to compare properties.}
    \vspace{-0.6cm}
\end{figure}

\noindent\textbf{Human Evaluation:} Due to the subjective nature of the VST task, a human evaluation is required. We randomly selected 150 image sequences from the test set and asked three MTurk annotators to rank them or assess them with other methods. We use the AREL as a baseline. Further, our method of generating coherent stories is tested using a slightly weaker variation of the method without directionality. Since the most recent baselines are not publicly available, we cannot compare them to recent approaches. We’ll share the selected sequences to aid future comparisons. 


In \figref{fig:dir_compare_qual}, we assess stories coherency. To begin, we examine the importance of modeling direction-aware interactions. In our comparison, we changed only one aspect of our model. We used the same factor for preceding and subsequent interactions. We show that on VIST metrics the effect is relatively small (\ie, no-direction; see \tabref{tab:direction_compare}). However, the human-evaluation comparison shows a significant coherency improvement (64.2\% \vs 28.7\%), which is not revealed with classical evaluation.  Also, a comparison against the AREL baseline demonstrates a more significant improvement (70.24\% \vs 25.32\%).  

In \figref{fig:comparative}  we provide the results when asking annotators to pick the most human-like story. We use the majority vote to decide the best model per story.  The generated stories outperform the AREL baseline (73.87\% \vs 22.53\%). Surprisingly, in many cases, the annotators found the generated stories to be more human-like than the ground truth stories (41\% \vs 48.57\%). 
\begin{table}[t]
 \centering
 \vspace{-0.3cm}
\caption{Human evaluation results for rating survey (scores are between 1-5).}
\vspace{-0.3cm}
 \scalebox{1}{
    \setlength{\tabcolsep}{1.0pt}
        \begin{tabular}{c|cccccc} 
            \toprule
            Method                    & Focused     & Coherent     & Share           & Human-like           & Grounded           & Detailed\\ 
            \midrule
            AREL                      & 3.49        & 3.18   & 3.18          & 3.26          & 3.32          & 3.15\\
            Ours           & 3.67        & 3.52           & 3.20 & 3.56         & 3.54           & 3.32\\
            GT           & \textbf{3.72}        & \textbf{3.57}            & \textbf{3.34} & \textbf{3.64}          & \textbf{3.56}            & \textbf{3.53}\\
            \bottomrule
        \end{tabular}
}
        \vspace{-0.3cm}
     \label{tab:rating}
\end{table}

To further evaluate the quality of the stories, we follow the criteria set by the Visual Storytelling Challenge\footnote{\url{http://visionandlanguage.net/workshop2018}} and conduct a survey where judges are asked to rate six categories between 1-5: 
\emph{1. Focused}: the story contains information that is ``naturally'' relevant to the rest of the story;
\emph{2. Coherence}: the sentences in the story are related and consistent;
\emph{3. Share}: the inclination to share the story; 
\emph{4. Human-like}: the story was likely written by a human; 
\emph{5. Grounded}: the story directly reflects concrete entities in the image; and 
\emph{6. Detailed}: the story provides an appropriate level of detail. 
To obtain the final score, we average the annotators' scores per sample, followed by averaging across the entire sample set. From \tabref{tab:rating} we observe: the model improved on all the criteria compared to the AREL model. Importantly, the generated stories are comparable to the ground-truth stories, indicating  success in reducing the shortcomings found in prior methods. 
Nonetheless, the  level of detail is still lacking, supporting the observation of \citet{holtzman2019curious} that current decoding strategies tend to generate well-formed yet somewhat generic text.

\begin{table}[t!]
    \centering
    \caption{Story generation ablation analysis.}
    \vspace{-0.4cm}
	\resizebox{\linewidth}{!}{%
    \begin{tabular}{cc|ccc}
    		\toprule
    		\multicolumn{2}{c}{ \textbf{Model}}       & \textbf{Text Rep. }      & \textbf{Sent. Rep.}           \\ 
    		\midrule
    		\multicolumn{2}{c}{AREL~\cite{Wang2018NoMA}}                   & 0.16     & 0.4       \\ 
    		\midrule
    		\multicolumn{1}{c}{\textbf{BOG prior}} & \textbf{Intra-repetition reg.} &	\multicolumn{2}{c}{} \\ 
    		\midrule
    		No                                 & No          & 0.14     & 0.33      \\
    		Yes                                & No          & 0.10     & 0.18      \\
    		No                                 & Yes         & 0.04     & 0.04      \\
    		Yes                                & Yes         & \textbf{0.008 }   & \textbf{0.0  }   \\
    		\bottomrule
    \end{tabular}
    }
    \label{tab:represults}
    \vspace{-0.3cm}
\end{table}
\begin{figure}[t]
	\includegraphics[width=0.65\linewidth]{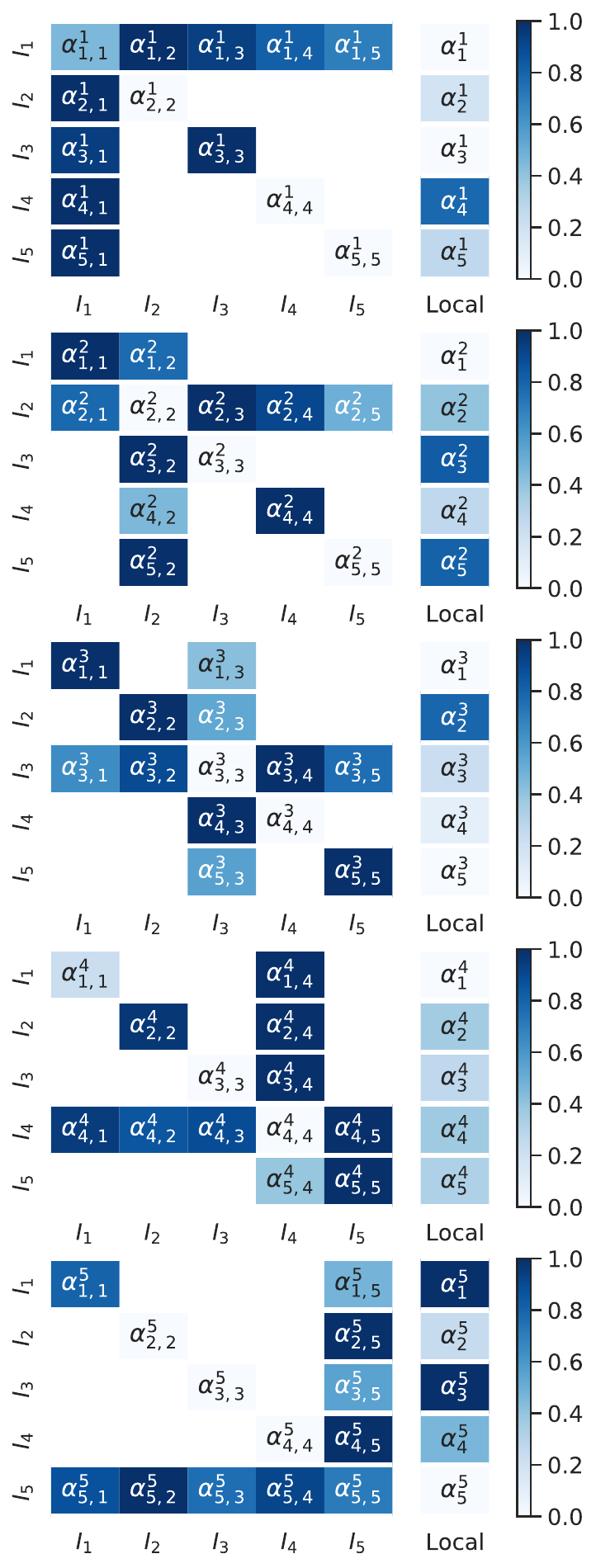}
	\vspace{-0.4cm}
	\caption{OIA scalar values (\ie, $\alpha_i^s$ and $\alpha_{i,s}^s$ in Sec.~3.1.1). The top map corresponds to the first sentence (\ie, $s=1$) and bottom one to the last sentence (\ie, $s=5$)}
	\label{fig:scalars_all}
	\vspace{-0.4cm}
\end{figure}


\noindent\textbf{Ablation study:} 
In Tab.~\ref{tab:direction_compare} we conduct an ablation study for two novel components in our model: 1) Attention: 
In `w/o attention,' we removed both OIA and ISA, which dropped the METEOR score  to 35.8\%. For the method referred to as `no-direction,' we use the same factor for preceding and subsequent interaction (\ie, $L_\text{bwd}=L_\text{fwd}$ and $R_\text{bwd}=R_\text{fwd}$). Here, METEOR results drop by 0.7\%. Hence, ordered interactions are beneficial. 2) We assess the decoding components (\secref{sec:decoder}). We first remove the intra-repetition regularization (Eq. \eqref{intra_rep_eq}), which causes METEOR score to drop by 0.6\%.  Removing the popular words count  ($\phi'_{s,t}$), results in a 0.4\% drop on METEOR. The METEOR score drops by 0.4\% when we remove the BOW prior. Last, we replace the GRU decoding layer with a Transformer, which did not change results a lot. 

In \tabref{tab:oia_interactions}  Further, we assess the necessity of different factors used in OIA. All factors contribute to the model's performance and the directional factors (\ie,
$\Psi_{\text{fwd}}$ and $\Psi_{\text{bwd}}$) have  the biggest impact.

In \tabref{tab:represults}, we show the ability to reduce repetitions. Text repetitiveness is measured by the repetition rate of non-singleton n-grams within each story. In our experiment, we use up to 4-grams. The use of intra-repetition regularization reduces text repetition (0.14 to 0.04). Combined with the trainable bag-of-words prior module, we further improve this measure (0.008 \vs 0.14). We also report sentence repetitiveness, \ie,  the average number of repeated sentences in a story.

\begin{figure}[ht]
	\centering
	\includegraphics[width=0.9\linewidth]{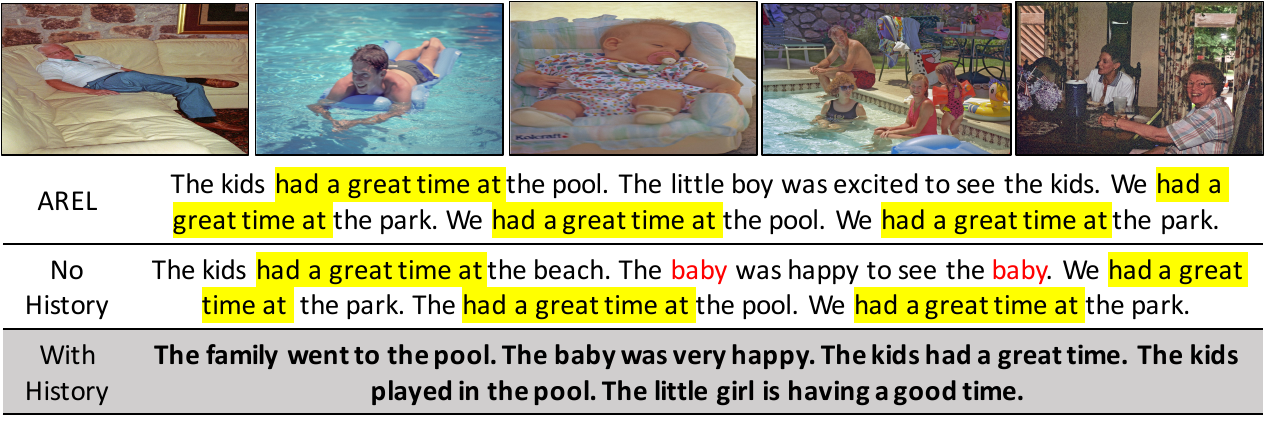}
    \vspace{-0.3cm}
	\caption{An illustration of an image sequence along with three different stories generated by: (1) AREL baseline~\cite{Wang2018NoMA}, (2) No History: a model without intra-repetition regularization and BOW prior (see \secref{sec:decoder}); 
	and (3) With History: the final model. Repeated sentences are highlighted with a  \hl{yellow colored marker}.  Repeated words in a sentence are emphasized in {\color{red}red} color.}
	\vspace{-0.3cm}
	\label{fig:repetition_coherance}
\end{figure}

\begin{figure}[t]
	\centering
	\includegraphics[width=0.9\linewidth]{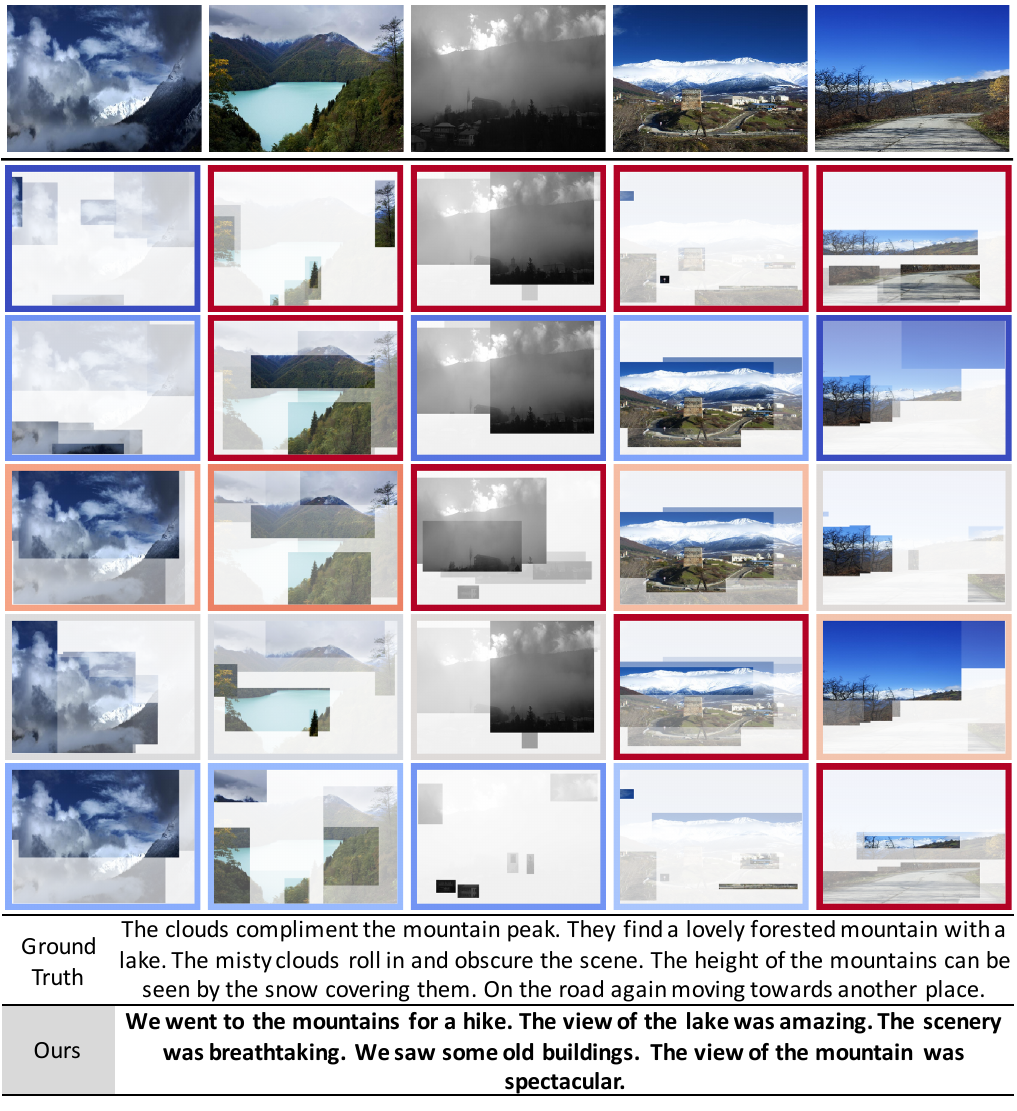}
\vspace{-0.4cm}
\caption{Illustration of OIA and ISA attention maps, the ground-truth story and the final generated story. Each row corresponds to a story sentence and shows objects OIA highlights. The attended images' border specifies the relevancy to sentence generation, from red (important) to blue (not important).}
\vspace{-0.5cm}
\label{fig:attention}
\end{figure}

In \figref{fig:scalars_all}, we illustrate for each sentence, the value of the importance calibration scalars (\ie, $\alpha_i^s$ and $\alpha_{i,s}^s$ in Eq.~\ref{eq:b_a},~\ref{eq:b_b}, and ~\ref{eq:b_c}).  Intuitively, these values indicate the importance of different image-to-image messages. We focus our analysis on the sentence-corresponding image  (\ie, $i=s$). We observe that the self-message scalars (\ie, $\mu_{i\to i}$) of the sentences in the middle of the sequence, \ie, sentences (2,3, and 4), are low. This indicates that the  images in the middle of the sequence rely more on the other images. The beginning and the ending of the story depend more on the local factors. Notably, the most substantial influence is given to the following image (\ie, $\alpha_{i,i+1}^i$). This means that while generating the current sentence, the OIA decision is based mostly on the next image. This is intuitive as it helps to advance the narrative in a desired direction.




\section{Qualitative Results}
\label{seq:qual}

In \figref{fig:attention} we illustrate the attention maps along with the generated story. The first sentence, ``We went to the mountains,'' sets the theme for the story, which requires the processing of subsequent images. Notably, the ISA module picked the subsequent images. In contrast, for the second sentence, the attention focuses mainly on the second image resulting in a description of the lake observed exclusively in this image.  The third sentence relates to the scenery. Hence the attention focuses on preceding and subsequent images.

In \figref{fig:repetition_coherance}, we show the ability of the method in reducing repetitions. We observe the AREL baseline to repeat the same sentences, for example, ``...had a great time at...''. We also observe this repetitiveness when we remove the bag-of-words prior and the intra-sentence regularization (\ie, No History column). Nevertheless, the method remains on topic,  \ie, family in the pool.

\section{Discussion}

The VIST dataset focuses primarily on one topic and relatively simple stories. Our method may limit longer stories with non-linear narratives. Annotating longer stories is significantly more challenging since even the tiniest fraction of five-image stories is already difficult to cover. Future work on visual storytelling may consider employing image-text models trained on hundreds of millions of image-text pairs, such as CLIP~\cite{radford2021learning}, to overcome the lack of annotated data.
\begin{table}[t]

		\centering
	\caption{Components ablation analysis.}
	\vspace{-0.4cm}
	%
		\begin{tabular}{cccccc}
			\toprule
			\textbf{Model}            & \textbf{M}        & \textbf{B-4}        &  \textbf{R}        & \textbf{C}        &  \textbf{\#Params}\\
			\midrule
					\hline\multicolumn{6}{c}{\textbf{attention}}    \\\hline 
			w/o attention              & 35.8          & 13.6  & 29.7 & 7.2  & 11M \\
			no-direction & 36.1          & 14.5  & 28.9  & 8.4        & 12M\\
								\hline\multicolumn{6}{c}{\textbf{decoding}}    \\\hline 

			w/o rep. regularization  & 36.2    & 14.5 & 29.8 & 8.7   & 13M\\
			w/o count norm & 36.2    & 14.6   & 29.9 & 9.4 & 13M\\
			w/o BOW prior  & 36.2    & 14.5   & 30.0 & 9.7  & 13M\\
			Transformer &   36.7 & \textbf{15.7} & 30.0 & 9.9  & 13M \\
			\midrule

			\textbf{Ours}  & \textbf{36.8} & 15.3  & \textbf{30.2}  & \textbf{10.1}    & 13M\\
			\bottomrule
	\end{tabular}
	\vspace{-.6cm}
	\label{tab:direction_compare}
\end{table}
\begin{table}[t]
    \centering
    \caption{Factor ablation analysis.}
    \vspace{-0.4cm}
        \begin{tabular}{ccccccc} 
    		\toprule
    		\textbf{Local}            & \textbf{Self}            & \textbf{Directional}            & \textbf{M}   & 
    		\textbf{B-4} &
    		    		\textbf{R}        & 
    		\textbf{C}\\   
    		\midrule
    		$\times$    & \checkmark    & \checkmark & 36.2      & 14.5 & 30.0  & 9.3    \\
    		\checkmark    & $\times$   & \checkmark    &  36.0       & 14.4 & 29.8          & 9.2  \\
    		\checkmark    & \checkmark    & $\times$    &   35.5     & 14.2 & 29.9          & 8.5  \\
    		\midrule
    		\checkmark    &
    		\checkmark    &
    		\checkmark    &
    		           \textbf{36.8} &           \textbf{15.3}        & \textbf{30.2}        & \textbf{10.1} \\
    		
    		\bottomrule
        \end{tabular}
    \label{tab:oia_interactions}
    \vspace{-.7cm}
\end{table}


\section{Conclusion}
We present a novel approach for VST, which encourages  coherency of generated  story. We incorporate structure between images with a new attention method that selects the important objects in an ordered image sequence. Human evaluation and quantitative analysis demonstrate that the approach outperforms existing methods. Further, we perform ablation analysis to show  effectiveness.

\begin{acks}
This work was partially supported by NSF under Grants 1718221, 2008387, 2045586, 2106825, MRI 1725729, NIFA award 2020-67021-32799, Israeli Science Foundation grant number 1390/19, Microsoft Corporation and AWS Cloud Credit for Research.
\end{acks}

\bibliographystyle{ACM-Reference-Format}
\balance 
\bibliography{acmmm}


\begin{thebibliography}{50}


\ifx \showCODEN    \undefined \def \showCODEN     #1{\unskip}     \fi
\ifx \showDOI      \undefined \def \showDOI       #1{#1}\fi
\ifx \showISBNx    \undefined \def \showISBNx     #1{\unskip}     \fi
\ifx \showISBNxiii \undefined \def \showISBNxiii  #1{\unskip}     \fi
\ifx \showISSN     \undefined \def \showISSN      #1{\unskip}     \fi
\ifx \showLCCN     \undefined \def \showLCCN      #1{\unskip}     \fi
\ifx \shownote     \undefined \def \shownote      #1{#1}          \fi
\ifx \showarticletitle \undefined \def \showarticletitle #1{#1}   \fi
\ifx \showURL      \undefined \def \showURL       {\relax}        \fi
\providecommand\bibfield[2]{#2}
\providecommand\bibinfo[2]{#2}
\providecommand\natexlab[1]{#1}
\providecommand\showeprint[2][]{arXiv:#2}

\bibitem[Anderson et~al\mbox{.}(2017)]%
        {Anderson2017BottomUpAT}
\bibfield{author}{\bibinfo{person}{Peter Anderson}, \bibinfo{person}{Xiaodong
  He}, \bibinfo{person}{Chris Buehler}, \bibinfo{person}{Damien Teney},
  \bibinfo{person}{Mark Johnson}, \bibinfo{person}{Stephen Gould}, {and}
  \bibinfo{person}{Lei Zhang}.} \bibinfo{year}{2017}\natexlab{}.
\newblock \showarticletitle{Bottom-Up and Top-Down Attention for Image
  Captioning and Visual Question Answering}. In
  \bibinfo{booktitle}{\emph{CVPR}}.
\newblock


\bibitem[Banerjee and Lavie(2005)]%
        {Banerjee2005METEORAA}
\bibfield{author}{\bibinfo{person}{Satanjeev Banerjee} {and}
  \bibinfo{person}{Alon Lavie}.} \bibinfo{year}{2005}\natexlab{}.
\newblock \showarticletitle{METEOR: An Automatic Metric for MT Evaluation with
  Improved Correlation with Human Judgments}. In
  \bibinfo{booktitle}{\emph{ACL}}.
\newblock


\bibitem[Barnard et~al\mbox{.}(2003)]%
        {Barnard2003MatchingWA}
\bibfield{author}{\bibinfo{person}{Kobus Barnard},
  \bibinfo{person}{Pinar~Duygulu Sahin}, \bibinfo{person}{David~A. Forsyth},
  \bibinfo{person}{Nando de Freitas}, \bibinfo{person}{David~M. Blei}, {and}
  \bibinfo{person}{Michael~I. Jordan}.} \bibinfo{year}{2003}\natexlab{}.
\newblock \showarticletitle{Matching Words and Pictures}.
\newblock \bibinfo{journal}{\emph{JMLR}} (\bibinfo{year}{2003}).
\newblock


\bibitem[Ben-Younes et~al\mbox{.}(2017)]%
        {ben2017mutan}
\bibfield{author}{\bibinfo{person}{Hedi Ben-Younes}, \bibinfo{person}{R{\'e}mi
  Cadene}, \bibinfo{person}{Matthieu Cord}, {and} \bibinfo{person}{Nicolas
  Thome}.} \bibinfo{year}{2017}\natexlab{}.
\newblock \showarticletitle{Mutan: Multimodal tucker fusion for visual question
  answering}. In \bibinfo{booktitle}{\emph{ICCV}}.
\newblock


\bibitem[Chen and Zitnick(2015)]%
        {Chen2015MindsEA}
\bibfield{author}{\bibinfo{person}{Xinlei Chen} {and}
  \bibinfo{person}{C.~Lawrence Zitnick}.} \bibinfo{year}{2015}\natexlab{}.
\newblock \showarticletitle{Mind's eye: A recurrent visual representation for
  image caption generation}. In \bibinfo{booktitle}{\emph{CVPR}}.
\newblock


\bibitem[Devlin et~al\mbox{.}(2019)]%
        {devlin2018bert}
\bibfield{author}{\bibinfo{person}{Jacob Devlin}, \bibinfo{person}{Ming-Wei
  Chang}, \bibinfo{person}{Kenton Lee}, {and} \bibinfo{person}{Kristina
  Toutanova}.} \bibinfo{year}{2019}\natexlab{}.
\newblock \showarticletitle{Bert: Pre-training of deep bidirectional
  transformers for language understanding}.
\newblock \bibinfo{journal}{\emph{NAACL}} (\bibinfo{year}{2019}).
\newblock


\bibitem[Dosovitskiy et~al\mbox{.}(2021)]%
        {dosovitskiy2021image}
\bibfield{author}{\bibinfo{person}{Alexey Dosovitskiy}, \bibinfo{person}{Lucas
  Beyer}, \bibinfo{person}{Alexander Kolesnikov}, \bibinfo{person}{Dirk
  Weissenborn}, \bibinfo{person}{Xiaohua Zhai}, \bibinfo{person}{Thomas
  Unterthiner}, \bibinfo{person}{Mostafa Dehghani}, \bibinfo{person}{Matthias
  Minderer}, \bibinfo{person}{Georg Heigold}, \bibinfo{person}{Sylvain Gelly},
  {et~al\mbox{.}}} \bibinfo{year}{2021}\natexlab{}.
\newblock \showarticletitle{An image is worth 16x16 words: Transformers for
  image recognition at scale}.
\newblock \bibinfo{journal}{\emph{ICLR}} (\bibinfo{year}{2021}).
\newblock


\bibitem[Fan et~al\mbox{.}(2021)]%
        {fan2021visual}
\bibfield{author}{\bibinfo{person}{Ruichao Fan}, \bibinfo{person}{Hanli Wang},
  \bibinfo{person}{Jinjing Gu}, {and} \bibinfo{person}{Xianhui Liu}.}
  \bibinfo{year}{2021}\natexlab{}.
\newblock \showarticletitle{Visual Storytelling with Hierarchical BERT Semantic
  Guidance}.
\newblock In \bibinfo{booktitle}{\emph{ACM Multimedia Asia}}.
  \bibinfo{pages}{1--7}.
\newblock


\bibitem[Fukui et~al\mbox{.}(2016)]%
        {fukui2016multimodal}
\bibfield{author}{\bibinfo{person}{Akira Fukui}, \bibinfo{person}{Dong~Huk
  Park}, \bibinfo{person}{Daylen Yang}, \bibinfo{person}{Anna Rohrbach},
  \bibinfo{person}{Trevor Darrell}, {and} \bibinfo{person}{Marcus Rohrbach}.}
  \bibinfo{year}{2016}\natexlab{}.
\newblock \showarticletitle{Multimodal compact bilinear pooling for visual
  question answering and visual grounding}. In
  \bibinfo{booktitle}{\emph{EMNLP}}.
\newblock


\bibitem[Gao et~al\mbox{.}(2019)]%
        {gao2019dynamic}
\bibfield{author}{\bibinfo{person}{Peng Gao}, \bibinfo{person}{Zhengkai Jiang},
  \bibinfo{person}{Haoxuan You}, \bibinfo{person}{Pan Lu},
  \bibinfo{person}{Steven~CH Hoi}, \bibinfo{person}{Xiaogang Wang}, {and}
  \bibinfo{person}{Hongsheng Li}.} \bibinfo{year}{2019}\natexlab{}.
\newblock \showarticletitle{Dynamic fusion with intra-and inter-modality
  attention flow for visual question answering}. In
  \bibinfo{booktitle}{\emph{CVPR}}.
\newblock


\bibitem[Gonzalez-Rico and Pineda(2018)]%
        {GonzalezRico2018ContextualizeSA}
\bibfield{author}{\bibinfo{person}{Diana Gonzalez-Rico} {and}
  \bibinfo{person}{Gibran~Fuentes Pineda}.} \bibinfo{year}{2018}\natexlab{}.
\newblock \showarticletitle{Contextualize, Show and Tell: A Neural Visual
  Storyteller}. In \bibinfo{booktitle}{\emph{Storytelling Workshop, NAACL}}.
\newblock


\bibitem[Guo et~al\mbox{.}(2020)]%
        {guo2020normalized}
\bibfield{author}{\bibinfo{person}{Longteng Guo}, \bibinfo{person}{Jing Liu},
  \bibinfo{person}{Xinxin Zhu}, \bibinfo{person}{Peng Yao},
  \bibinfo{person}{Shichen Lu}, {and} \bibinfo{person}{Hanqing Lu}.}
  \bibinfo{year}{2020}\natexlab{}.
\newblock \showarticletitle{Normalized and geometry-aware self-attention
  network for image captioning}. In \bibinfo{booktitle}{\emph{CVPR}}.
\newblock


\bibitem[He et~al\mbox{.}(2015)]%
        {He2015DeepRL}
\bibfield{author}{\bibinfo{person}{Kaiming He}, \bibinfo{person}{Xiangyu
  Zhang}, \bibinfo{person}{Shaoqing Ren}, {and} \bibinfo{person}{Jian Sun}.}
  \bibinfo{year}{2015}\natexlab{}.
\newblock \showarticletitle{Deep Residual Learning for Image Recognition}. In
  \bibinfo{booktitle}{\emph{CVPR}}.
\newblock


\bibitem[Holtzman et~al\mbox{.}(2020)]%
        {holtzman2019curious}
\bibfield{author}{\bibinfo{person}{Ari Holtzman}, \bibinfo{person}{Jan Buys},
  \bibinfo{person}{Li Du}, \bibinfo{person}{Maxwell Forbes}, {and}
  \bibinfo{person}{Yejin Choi}.} \bibinfo{year}{2020}\natexlab{}.
\newblock \showarticletitle{The curious case of neural text degeneration}. In
  \bibinfo{booktitle}{\emph{ICLR}}.
\newblock


\bibitem[Hong et~al\mbox{.}(2020)]%
        {hong2020diverse}
\bibfield{author}{\bibinfo{person}{Xudong Hong}, \bibinfo{person}{Rakshith
  Shetty}, \bibinfo{person}{Asad Sayeed}, \bibinfo{person}{Khushboo Mehra},
  \bibinfo{person}{Vera Demberg}, {and} \bibinfo{person}{Bernt Schiele}.}
  \bibinfo{year}{2020}\natexlab{}.
\newblock \showarticletitle{Diverse and Relevant Visual Storytelling with Scene
  Graph Embeddings}. In \bibinfo{booktitle}{\emph{CONLL}}.
\newblock


\bibitem[Hsu et~al\mbox{.}(2020)]%
        {hsu2020knowledge}
\bibfield{author}{\bibinfo{person}{Chao-Chun Hsu}, \bibinfo{person}{Zi-Yuan
  Chen}, \bibinfo{person}{Chi-Yang Hsu}, \bibinfo{person}{Chih-Chia Li},
  \bibinfo{person}{Tzu-Yuan Lin}, \bibinfo{person}{Ting-Hao Huang}, {and}
  \bibinfo{person}{Lun-Wei Ku}.} \bibinfo{year}{2020}\natexlab{}.
\newblock \showarticletitle{Knowledge-Enriched Visual Storytelling}. In
  \bibinfo{booktitle}{\emph{AAAI}}.
\newblock


\bibitem[Huang et~al\mbox{.}(2018)]%
        {Huang2018HierarchicallySR}
\bibfield{author}{\bibinfo{person}{Qiuyuan Huang}, \bibinfo{person}{Zhe Gan},
  \bibinfo{person}{Asli Çelikyilmaz}, \bibinfo{person}{Dapeng Wu},
  \bibinfo{person}{Jianfeng Wang}, {and} \bibinfo{person}{Xiaodong He}.}
  \bibinfo{year}{2018}\natexlab{}.
\newblock \showarticletitle{Hierarchically Structured Reinforcement Learning
  for Topically Coherent Visual Story Generation}. In
  \bibinfo{booktitle}{\emph{AAAI}}.
\newblock


\bibitem[Huang et~al\mbox{.}(2016)]%
        {Huang2016VisualS}
\bibfield{author}{\bibinfo{person}{Ting-Hao Huang}, \bibinfo{person}{Francis
  Ferraro}, \bibinfo{person}{Nasrin Mostafazadeh}, \bibinfo{person}{Ishan
  Misra}, \bibinfo{person}{Aishwarya Agrawal}, \bibinfo{person}{Jacob Devlin},
  \bibinfo{person}{Ross~B. Girshick}, \bibinfo{person}{Xiaodong He},
  \bibinfo{person}{Pushmeet Kohli}, \bibinfo{person}{Dhruv Batra},
  \bibinfo{person}{C.~Lawrence Zitnick}, \bibinfo{person}{Devi Parikh},
  \bibinfo{person}{Lucy Vanderwende}, \bibinfo{person}{Michel Galley}, {and}
  \bibinfo{person}{Margaret Mitchell}.} \bibinfo{year}{2016}\natexlab{}.
\newblock \showarticletitle{Visual Storytelling}. In
  \bibinfo{booktitle}{\emph{NAACL}}.
\newblock


\bibitem[Kim et~al\mbox{.}(2018b)]%
        {kim2018bilinear}
\bibfield{author}{\bibinfo{person}{Jin-Hwa Kim}, \bibinfo{person}{Jaehyun Jun},
  {and} \bibinfo{person}{Byoung-Tak Zhang}.} \bibinfo{year}{2018}\natexlab{b}.
\newblock \showarticletitle{Bilinear attention networks}. In
  \bibinfo{booktitle}{\emph{NeurIPS}}.
\newblock


\bibitem[Kim et~al\mbox{.}(2017)]%
        {kim2016hadamard}
\bibfield{author}{\bibinfo{person}{Jin-Hwa Kim}, \bibinfo{person}{Kyoung-Woon
  On}, \bibinfo{person}{Woosang Lim}, \bibinfo{person}{Jeonghee Kim},
  \bibinfo{person}{Jung-Woo Ha}, {and} \bibinfo{person}{Byoung-Tak Zhang}.}
  \bibinfo{year}{2017}\natexlab{}.
\newblock \showarticletitle{Hadamard product for low-rank bilinear pooling}. In
  \bibinfo{booktitle}{\emph{ICLR}}.
\newblock


\bibitem[Kim et~al\mbox{.}(2018a)]%
        {Kim2018GLAC}
\bibfield{author}{\bibinfo{person}{Taehyeong Kim}, \bibinfo{person}{Min-Oh
  Heo}, \bibinfo{person}{Seonil Son}, \bibinfo{person}{Kyoung-Wha Park}, {and}
  \bibinfo{person}{Byoung-Tak Zhang}.} \bibinfo{year}{2018}\natexlab{a}.
\newblock \showarticletitle{GLAC Net: GLocal Attention Cascading Networks for
  Multi-image Cued Story Generation}. In \bibinfo{booktitle}{\emph{CoRR}}.
\newblock


\bibitem[Li et~al\mbox{.}(2019)]%
        {Li2019InformativeVS}
\bibfield{author}{\bibinfo{person}{Jiacheng Li}, \bibinfo{person}{Haizhou Shi},
  \bibinfo{person}{Siliang Tang}, \bibinfo{person}{Fei Wu}, {and}
  \bibinfo{person}{Yueting Zhuang}.} \bibinfo{year}{2019}\natexlab{}.
\newblock \showarticletitle{Informative Visual Storytelling with Cross-modal
  Rules}. In \bibinfo{booktitle}{\emph{MM}}.
\newblock


\bibitem[Li et~al\mbox{.}(2020)]%
        {li2020oscar}
\bibfield{author}{\bibinfo{person}{Xiujun Li}, \bibinfo{person}{Xi Yin},
  \bibinfo{person}{Chunyuan Li}, \bibinfo{person}{Pengchuan Zhang},
  \bibinfo{person}{Xiaowei Hu}, \bibinfo{person}{Lei Zhang},
  \bibinfo{person}{Lijuan Wang}, \bibinfo{person}{Houdong Hu},
  \bibinfo{person}{Li Dong}, \bibinfo{person}{Furu Wei}, {et~al\mbox{.}}}
  \bibinfo{year}{2020}\natexlab{}.
\newblock \showarticletitle{Oscar: Object-semantics aligned pre-training for
  vision-language tasks}. In \bibinfo{booktitle}{\emph{ECCV}}.
\newblock


\bibitem[Lin(2004)]%
        {Lin2004ROUGEAP}
\bibfield{author}{\bibinfo{person}{Chin-Yew Lin}.}
  \bibinfo{year}{2004}\natexlab{}.
\newblock \showarticletitle{ROUGE: A Package For Automatic Evaluation Of
  Summaries}. In \bibinfo{booktitle}{\emph{ACL}}.
\newblock


\bibitem[Lu et~al\mbox{.}(2016)]%
        {lu2016hierarchical}
\bibfield{author}{\bibinfo{person}{Jiasen Lu}, \bibinfo{person}{Jianwei Yang},
  \bibinfo{person}{Dhruv Batra}, {and} \bibinfo{person}{Devi Parikh}.}
  \bibinfo{year}{2016}\natexlab{}.
\newblock \showarticletitle{Hierarchical question-image co-attention for visual
  question answering}. In \bibinfo{booktitle}{\emph{NeurIPS}}.
\newblock


\bibitem[Modi and Parde(2019)]%
        {Modi2019TheSR}
\bibfield{author}{\bibinfo{person}{Yatri Modi} {and} \bibinfo{person}{Natalie
  Parde}.} \bibinfo{year}{2019}\natexlab{}.
\newblock \showarticletitle{The Steep Road to Happily Ever after: an Analysis
  of Current Visual Storytelling Models}. In \bibinfo{booktitle}{\emph{Workshop
  on Shortcomings in Vision and Language, NAACL}}.
\newblock


\bibitem[Pan et~al\mbox{.}(2020)]%
        {pan2020x}
\bibfield{author}{\bibinfo{person}{Yingwei Pan}, \bibinfo{person}{Ting Yao},
  \bibinfo{person}{Yehao Li}, {and} \bibinfo{person}{Tao Mei}.}
  \bibinfo{year}{2020}\natexlab{}.
\newblock \showarticletitle{X-linear attention networks for image captioning}.
  In \bibinfo{booktitle}{\emph{CVPR}}.
\newblock


\bibitem[Papineni et~al\mbox{.}(2001)]%
        {Papineni2001BleuAM}
\bibfield{author}{\bibinfo{person}{Kishore Papineni}, \bibinfo{person}{Salim
  Roukos}, \bibinfo{person}{Todd Ward}, {and} \bibinfo{person}{Wei-Jing Zhu}.}
  \bibinfo{year}{2001}\natexlab{}.
\newblock \showarticletitle{Bleu: a Method for Automatic Evaluation of Machine
  Translation}. In \bibinfo{booktitle}{\emph{ACL}}.
\newblock


\bibitem[Park and Kim(2015)]%
        {Park2015ExpressingAI}
\bibfield{author}{\bibinfo{person}{Cesc~C. Park} {and} \bibinfo{person}{Gunhee
  Kim}.} \bibinfo{year}{2015}\natexlab{}.
\newblock \showarticletitle{Expressing an Image Stream with a Sequence of
  Natural Sentences}. In \bibinfo{booktitle}{\emph{NeurIPS}}.
\newblock


\bibitem[Qi et~al\mbox{.}(2021)]%
        {qi2021latent}
\bibfield{author}{\bibinfo{person}{Mengshi Qi}, \bibinfo{person}{Jie Qin},
  \bibinfo{person}{Di Huang}, \bibinfo{person}{Zhiqiang Shen},
  \bibinfo{person}{Yi Yang}, {and} \bibinfo{person}{Jiebo Luo}.}
  \bibinfo{year}{2021}\natexlab{}.
\newblock \showarticletitle{Latent Memory-augmented Graph Transformer for
  Visual Storytelling}. In \bibinfo{booktitle}{\emph{Proceedings of the 29th
  ACM International Conference on Multimedia}}. \bibinfo{pages}{4892--4901}.
\newblock


\bibitem[Radford et~al\mbox{.}(2021)]%
        {radford2021learning}
\bibfield{author}{\bibinfo{person}{Alec Radford}, \bibinfo{person}{Jong~Wook
  Kim}, \bibinfo{person}{Chris Hallacy}, \bibinfo{person}{Aditya Ramesh},
  \bibinfo{person}{Gabriel Goh}, \bibinfo{person}{Sandhini Agarwal},
  \bibinfo{person}{Girish Sastry}, \bibinfo{person}{Amanda Askell},
  \bibinfo{person}{Pamela Mishkin}, \bibinfo{person}{Jack Clark},
  {et~al\mbox{.}}} \bibinfo{year}{2021}\natexlab{}.
\newblock \showarticletitle{Learning transferable visual models from natural
  language supervision}. In \bibinfo{booktitle}{\emph{ICML}}.
\newblock


\bibitem[Schwartz et~al\mbox{.}(2017)]%
        {Schwartz2017HighOrderAM}
\bibfield{author}{\bibinfo{person}{Idan Schwartz},
  \bibinfo{person}{Alexander~G. Schwing}, {and} \bibinfo{person}{Tamir Hazan}.}
  \bibinfo{year}{2017}\natexlab{}.
\newblock \showarticletitle{High-Order Attention Models for Visual Question
  Answering}. In \bibinfo{booktitle}{\emph{NeurIPS}}.
\newblock


\bibitem[Schwartz et~al\mbox{.}(2019)]%
        {Schwartz2019FactorGA}
\bibfield{author}{\bibinfo{person}{Idan Schwartz}, \bibinfo{person}{Seunghak
  Yu}, \bibinfo{person}{Tamir Hazan}, {and} \bibinfo{person}{Alexander~G
  Schwing}.} \bibinfo{year}{2019}\natexlab{}.
\newblock \showarticletitle{Factor graph attention}. In
  \bibinfo{booktitle}{\emph{CVPR}}.
\newblock


\bibitem[Su et~al\mbox{.}(2021)]%
        {su2021bert}
\bibfield{author}{\bibinfo{person}{Jing Su}, \bibinfo{person}{Qingyun Dai},
  \bibinfo{person}{Frank Guerin}, {and} \bibinfo{person}{Mian Zhou}.}
  \bibinfo{year}{2021}\natexlab{}.
\newblock \showarticletitle{BERT-hLSTMs: BERT and hierarchical LSTMs for visual
  storytelling}.
\newblock \bibinfo{journal}{\emph{Computer Speech \& Language}}
  \bibinfo{volume}{67} (\bibinfo{year}{2021}), \bibinfo{pages}{101169}.
\newblock


\bibitem[Sutskever et~al\mbox{.}(2014)]%
        {Sutskever2014SequenceTS}
\bibfield{author}{\bibinfo{person}{Ilya Sutskever}, \bibinfo{person}{Oriol
  Vinyals}, {and} \bibinfo{person}{Quoc~V. Le}.}
  \bibinfo{year}{2014}\natexlab{}.
\newblock \showarticletitle{Sequence to Sequence Learning with Neural
  Networks}. In \bibinfo{booktitle}{\emph{NeurIPS}}.
\newblock


\bibitem[Szegedy et~al\mbox{.}(2015)]%
        {Szegedy2015RethinkingTI}
\bibfield{author}{\bibinfo{person}{Christian Szegedy}, \bibinfo{person}{Vincent
  Vanhoucke}, \bibinfo{person}{Sergey Ioffe}, \bibinfo{person}{Jonathon
  Shlens}, {and} \bibinfo{person}{Zbigniew Wojna}.}
  \bibinfo{year}{2015}\natexlab{}.
\newblock \showarticletitle{Rethinking the Inception Architecture for Computer
  Vision}. In \bibinfo{booktitle}{\emph{CVPR}}.
\newblock


\bibitem[Tewel et~al\mbox{.}(2022)]%
        {tewel2021zero}
\bibfield{author}{\bibinfo{person}{Yoad Tewel}, \bibinfo{person}{Yoav Shalev},
  \bibinfo{person}{Idan Schwartz}, {and} \bibinfo{person}{Lior Wolf}.}
  \bibinfo{year}{2022}\natexlab{}.
\newblock \showarticletitle{Zero-Shot Image-to-Text Generation for
  Visual-Semantic Arithmetic}.
\newblock \bibinfo{journal}{\emph{CVPR}} (\bibinfo{year}{2022}).
\newblock


\bibitem[Vaswani et~al\mbox{.}(2017)]%
        {vaswani2017attention}
\bibfield{author}{\bibinfo{person}{Ashish Vaswani}, \bibinfo{person}{Noam
  Shazeer}, \bibinfo{person}{Niki Parmar}, \bibinfo{person}{Jakob Uszkoreit},
  \bibinfo{person}{Llion Jones}, \bibinfo{person}{Aidan~N Gomez},
  \bibinfo{person}{{\L}ukasz Kaiser}, {and} \bibinfo{person}{Illia
  Polosukhin}.} \bibinfo{year}{2017}\natexlab{}.
\newblock \showarticletitle{Attention is all you need}. In
  \bibinfo{booktitle}{\emph{NeurIPS}}.
\newblock


\bibitem[Vedantam et~al\mbox{.}(2014)]%
        {Vedantam2014CIDErCI}
\bibfield{author}{\bibinfo{person}{Ramakrishna Vedantam},
  \bibinfo{person}{C.~Lawrence Zitnick}, {and} \bibinfo{person}{Devi Parikh}.}
  \bibinfo{year}{2014}\natexlab{}.
\newblock \showarticletitle{CIDEr: Consensus-based image description
  evaluation}. In \bibinfo{booktitle}{\emph{CVPR}}.
\newblock


\bibitem[Vinyals et~al\mbox{.}(2014)]%
        {Vinyals2014ShowAT}
\bibfield{author}{\bibinfo{person}{Oriol Vinyals}, \bibinfo{person}{Alexander
  Toshev}, \bibinfo{person}{Samy Bengio}, {and} \bibinfo{person}{Dumitru
  Erhan}.} \bibinfo{year}{2014}\natexlab{}.
\newblock \showarticletitle{Show and tell: A neural image caption generator}.
  In \bibinfo{booktitle}{\emph{CVPR}}.
\newblock


\bibitem[Wang et~al\mbox{.}(2019)]%
        {Wang2019StorytellingFA}
\bibfield{author}{\bibinfo{person}{Ruize Wang}, \bibinfo{person}{Zhongyu Wei},
  \bibinfo{person}{Piji Li}, \bibinfo{person}{Qi Zhang}, {and}
  \bibinfo{person}{Xuanjing Huang}.} \bibinfo{year}{2019}\natexlab{}.
\newblock \showarticletitle{Storytelling from an Image Stream Using Scene
  Graphs}. In \bibinfo{booktitle}{\emph{AAAI}}.
\newblock


\bibitem[Wang et~al\mbox{.}(2018)]%
        {Wang2018NoMA}
\bibfield{author}{\bibinfo{person}{Xin Wang}, \bibinfo{person}{Wenhu Chen},
  \bibinfo{person}{Yuan fang Wang}, {and} \bibinfo{person}{William~Yang Wang}.}
  \bibinfo{year}{2018}\natexlab{}.
\newblock \showarticletitle{No Metrics Are Perfect: Adversarial Reward Learning
  for Visual Storytelling}. In \bibinfo{booktitle}{\emph{ACL}}.
\newblock


\bibitem[Xu and Saenko(2016)]%
        {xu2016ask}
\bibfield{author}{\bibinfo{person}{Huijuan Xu} {and} \bibinfo{person}{Kate
  Saenko}.} \bibinfo{year}{2016}\natexlab{}.
\newblock \showarticletitle{Ask, attend and answer: Exploring question-guided
  spatial attention for visual question answering}. In
  \bibinfo{booktitle}{\emph{ECCV}}.
\newblock


\bibitem[Xu et~al\mbox{.}(2015)]%
        {Xu2015ShowAA}
\bibfield{author}{\bibinfo{person}{Kelvin Xu}, \bibinfo{person}{Jimmy Ba},
  \bibinfo{person}{Ryan Kiros}, \bibinfo{person}{Kyunghyun Cho},
  \bibinfo{person}{Aaron~C. Courville}, \bibinfo{person}{Ruslan Salakhutdinov},
  \bibinfo{person}{Richard~S. Zemel}, {and} \bibinfo{person}{Yoshua Bengio}.}
  \bibinfo{year}{2015}\natexlab{}.
\newblock \showarticletitle{Show, Attend and Tell: Neural Image Caption
  Generation with Visual Attention}. In \bibinfo{booktitle}{\emph{ICML}}.
\newblock


\bibitem[Yang et~al\mbox{.}(2019)]%
        {Yang2019KnowledgeableSA}
\bibfield{author}{\bibinfo{person}{Pengcheng Yang}, \bibinfo{person}{Fuli Luo},
  \bibinfo{person}{Peng Chen}, \bibinfo{person}{Lei Li}, \bibinfo{person}{Zhiyi
  Yin}, \bibinfo{person}{Xiaodong He}, {and} \bibinfo{person}{Xu Sun}.}
  \bibinfo{year}{2019}\natexlab{}.
\newblock \showarticletitle{Knowledgeable Storyteller: A Commonsense-Driven
  Generative Model for Visual Storytelling}. In
  \bibinfo{booktitle}{\emph{IJCAI}}.
\newblock


\bibitem[Yang et~al\mbox{.}(2015)]%
        {yang2015stacked}
\bibfield{author}{\bibinfo{person}{Zichao Yang}, \bibinfo{person}{Xiaodong He},
  \bibinfo{person}{Jianfeng Gao}, \bibinfo{person}{Li Deng}, {and}
  \bibinfo{person}{Alexander~J Smola}.} \bibinfo{year}{2015}\natexlab{}.
\newblock \bibinfo{title}{Stacked attention networks for image question
  answering}.
\newblock
\newblock


\bibitem[Yu et~al\mbox{.}(2021)]%
        {yu2021transitional}
\bibfield{author}{\bibinfo{person}{Youngjae Yu}, \bibinfo{person}{Jiwan Chung},
  \bibinfo{person}{Heeseung Yun}, \bibinfo{person}{Jongseok Kim}, {and}
  \bibinfo{person}{Gunhee Kim}.} \bibinfo{year}{2021}\natexlab{}.
\newblock \showarticletitle{Transitional Adaptation of Pretrained Models for
  Visual Storytelling}. In \bibinfo{booktitle}{\emph{CVPR}}.
\newblock


\bibitem[Yu et~al\mbox{.}(2018)]%
        {yu2018beyond}
\bibfield{author}{\bibinfo{person}{Zhou Yu}, \bibinfo{person}{Jun Yu},
  \bibinfo{person}{Chenchao Xiang}, \bibinfo{person}{Jianping Fan}, {and}
  \bibinfo{person}{Dacheng Tao}.} \bibinfo{year}{2018}\natexlab{}.
\newblock \showarticletitle{Beyond bilinear: Generalized multimodal factorized
  high-order pooling for visual question answering}. In
  \bibinfo{booktitle}{\emph{NeurIPS}}.
\newblock


\bibitem[Zhang et~al\mbox{.}(2020)]%
        {Zhang2020VisualSV}
\bibfield{author}{\bibinfo{person}{Bowen Zhang}, \bibinfo{person}{Hexiang Hu},
  {and} \bibinfo{person}{Fei Sha}.} \bibinfo{year}{2020}\natexlab{}.
\newblock \showarticletitle{Visual Storytelling via Predicting Anchor Word
  Embeddings in the Stories}. In \bibinfo{booktitle}{\emph{ICCV}}. Workshop on
  Closing the Loop Between Vision and Language.
\newblock


\bibitem[Zhang et~al\mbox{.}(2021)]%
        {zhang2021vinvl}
\bibfield{author}{\bibinfo{person}{Pengchuan Zhang}, \bibinfo{person}{Xiujun
  Li}, \bibinfo{person}{Xiaowei Hu}, \bibinfo{person}{Jianwei Yang},
  \bibinfo{person}{Lei Zhang}, \bibinfo{person}{Lijuan Wang},
  \bibinfo{person}{Yejin Choi}, {and} \bibinfo{person}{Jianfeng Gao}.}
  \bibinfo{year}{2021}\natexlab{}.
\newblock \showarticletitle{Vinvl: Revisiting visual representations in
  vision-language models}. In \bibinfo{booktitle}{\emph{CVPR}}.
\newblock


\end{thebibliography}

\end{document}